\documentclass[letterpaper, 10 pt, conference]{ieeeconf}  

\IEEEoverridecommandlockouts                              

\usepackage{ dsfont } 
\usepackage{bm} 
\usepackage{float}
\usepackage[cmex10]{amsmath}
\usepackage{algorithmic}
\usepackage{array}
\usepackage{mdwmath}
\usepackage{mdwtab}
\usepackage{eqparbox}
\usepackage{fixltx2e}
\usepackage{stfloats}
\usepackage{url}
\usepackage{amsmath} 
\usepackage{amssymb}  
\usepackage{amsfonts}
\usepackage{verbatim}
\usepackage{amsbsy}
\usepackage{graphicx}
\usepackage{subfig}
\usepackage[style=base]{caption}
\usepackage{epstopdf}
\usepackage{booktabs, multicol, multirow}
\usepackage{multicol}
\usepackage{bbm}
\usepackage{color}
\usepackage{makecell}

\usepackage[noadjust]{cite}
\usepackage[export]{adjustbox}

\usepackage{mathtools}

\usepackage[english]{babel}

\usepackage{xspace}

\DeclareRobustCommand{\ie}{i.e.\@\xspace}

\makeatletter
\DeclareRobustCommand{\etc}{%
    \@ifnextchar{.}%
        {etc}%
        {etc.\@\xspace}%
}

\newtheorem{theorem}{Theorem}
\newtheorem{corollary}{Corollary}

\renewcommand{\P}[1]{P \big ( #1 \big )}
\newcommand{\Cgamma}{ \bar{\mathcal{C}}_{\gamma} }
\newcommand{\Ceta}{ \bar{\mathcal{C}}_{\eta} }
\newcommand{\loss}[1]{\big ( #1 \big )}
\newcommand{\I}[1]{\mathbf{I} \big ( #1 \big )}
\newcommand{\W}[1]{W \big ( #1 \big )}
\newcommand{\U}[1]{U \big ( #1 \big )}
\newcommand{\V}[1]{V \big ( #1 \big )}
\newcommand{\E}[1]{\mathbb{E} \Big [  #1 \Big ]}
\newcommand{\Es}[1]{\mathbb{E} \big [  #1 \big ]}

\renewcommand{\H}[1]{H \big ( #1 \big )}

\title{\LARGE \bf
Environmental Information Improves Robotic Search Performance
}

\author{Harun Yetkin, Collin Lutz and Daniel J. Stilwell
\thanks{Harun Yetkin, Collin Lutz and Daniel J. Stilwell are with the Bradley Department of Electrical and Computer Engineering, Virginia Tech, Blacksburg, VA, USA }%
\thanks{contact email: yetkinh@vt.edu, stilwell@vt.edu}
}

\begin{document}

\maketitle
\thispagestyle{empty}
\pagestyle{empty}

\begin{abstract}

We address the problem where a mobile search agent seeks to find an unknown number of stationary objects distributed in a bounded search domain, and the search mission is subject to time/distance constraint. Our work accounts for false positives, false negatives and environmental uncertainty. We consider the case that the performance of a search sensor is dependent on the environment (e.g., clutter density), and therefore sensor performance is better in some locations than in others. We specifically consider applications where environmental information can be acquired either by a separate vehicle or by the same vehicle that performs the search task. Our main contribution in this study is to formally derive a decision-theoretic cost function to compute the locations where the environmental information should be acquired. For the cases where computing the optimal locations to sample the environment is computationally expensive, we offer an approximation approach that yields provable near-optimal paths. We show that our decision-theoretic cost function outperforms the information-maximization approach, which is often employed in similar applications.  

\end{abstract}

\section{Introduction}

We address search applications where a robotic system is to find an unknown number of objects in a bounded environment and within bounded time.  We assume that the environment affects sensor performance, and that variation in the environment throughout the search area causes search performance in some locations to be better than in other locations. Our prior work in~\cite{yetkin.etal.oceans2015} shows how guidance algorithms for search missions can incorporate stochastic knowledge of the environment to improve search performance. In this study, we specifically consider the cases where environmental information can be acquired as part of the overall search task. The principal contributions of this study show where environmental information should be acquired in order to improve overall search performance. We address two cases: (1) environmental characterization is performed prior to a search mission by a separate asset than the search vehicle, and (2) environmental characterization is performed at the same time as search by the same vehicle that performs the search task. In this study, we extend our findings in our prior work~\cite{yetkin.etal.oceans2015, yetkin.etal.oceans2016}.

We use a decision-theoretic value function that is associated with the accuracy of our estimate of the number of objects in the environment. Because search performance is dependent on the environment, knowledge of the environment can improve search performance due to better search plans. For example, one may choose to avoid searching areas that are known to contain excessive clutter and many false positives in favor of environments with few false positives.  In situations where the environment is poorly known, efforts to acquire environmental information may lead to improved search effectiveness.  We address the case that stochastic knowledge of the environment can be acquired, and we describe where the environment should be surveyed in order to improve overall search performance. One approach for selecting where to acquire environmental information is simply to characterize locations that yield the greatest reduction of uncertainty about the environment. In other words, one might seek to maximize reduction in entropy of the distribution that describe the environment, which is often employed in similar applications \cite{coleman2007nonlinear, papadimitriou2000entropy, elfes1992dynamic}. In contrast, we show that reducing uncertainty in the environment is not the best approach. A primary contribution of this work is to show that environmental information should be acquired at the locations where the greatest reduction of uncertainty in anticipated search performance will occur, where we define search performance as the probability that the estimate for the number of objects in the environment is correct. Computing the optimal locations to acquire the environment information can be computationally expensive when the planning horizon is large. To address the computational challenge of our approach, we show an approximation approach that yields provably near-optimal paths. 
 
The remainder of this paper is organized as follows. An overview of search theory and the benefit of acquiring environmental information in some search missions is provided in Section~\ref{sec:previousWork}. In Section~\ref{sec:search}, we formulate the search problem and define the observation model. In Section~\ref{sec:searcher}, we define the objective function that maximizes the estimation accuracy. In Section~\ref{sec:characterization} and Section~\ref{sec:compound}, we describe our proposed cost function to compute the locations where environmental characterization should be performed. Section~\ref{sec:numericalResults} provides the numerical results that illustrate our approach.

\section{Related Work}
\label{sec:previousWork}

\subsection{Search Theory}

Search theory is concerned with finding an optimal allocation of available search effort to locate a lost or hidden target, such that a reward specified as a measure of search effectiveness is maximized. Bernard Koopman offered one of the first systematic approaches to the search problem~\cite{Koopman1957}. His work laid the foundations of the problem and since then the search problem received a lot of attention, mainly from the operations research community. Some notable example include~\cite{Stone1972,Richardson1988,Kadane1971,Kimeldorf1979,Kress2008,Chung2012}. 

In a realistic search problem, there are certain limitations on performing search at a location such as imperfect sensor measurements or uncertain knowledge of the environment at that location. Noisy sensor measurements often include missed detections, \ie failing to detect the object, and false alarms, \ie generating a false detection event. Local environmental conditions can affect the number of missed detections and false alarms the sensor observes. Although there is an extensive literature on search theory (see e.g.,~\cite{benkoski1991survey, chung.etal.AR2011}), the issue of false alarms is seldom addressed.  Exceptions include~\cite{Pollock1964, Stone1972, Dobbie1973, Kress2008, chung2012analysis, kriheli2016optimal}, but  uncertainty in environmental conditions is not accounted for in these studies. In this work, the objective function we propose for assessing the value of searching a location builds upon prior work by accounting for multiple targets, false alarms, non-zero cost of moving to another search location, and uncertainty in the environment. To best of our knowledge, this is the first study to account for all these factors together. 

We do not address coverage problems as in~\cite{choset.AMAI2001} or exhaustive search.  Rather, we consider applications where there is a time or distance constraint, and we seek solutions where we achieve the best search performance within a time or distance constraint. Due to this constraint, optimal search paths may not visit every location. Indeed, in some scenarios it is possible that some locations are visited more than once while other locations are never visited at all. 

\subsection{Effect of Environment Information}

We consider that the local environmental conditions affect the sensor performance, and that the environment varies throughout the search region. Hence, the sensor performs better in some locations than in others. The environment at a location may not be known with certainty and we may only have a probabilistic knowledge on the environment. When possible, acquiring information about the local environmental conditions can significantly improve the results of a follow-on search mission. The question is, where to sample the environment so that the most improvement in search performance can be obtained. This is the fundamental question we seek to answer in this study, and we show that a new approach is needed to address the cases where the limited planning horizon of the mission does not allow exhaustive search of the environment.

The effect of the environment on search performance is well-known.  In subsea applications where sonar is used for search, variations in the seabed induce significant variation in probability of detection and probability of false alarm \cite{elmore.etal.OCEANS2007, zare.cobb.OCEANS.2013}. For terrestrial applications using ground penetrating radar, search effectiveness is dependent on background clutter and soil properties \cite{takahashi.etal.IEEESTAEORS.2011, takahashi.etal.JAG.2011, gader.etal.IEEEGRS.2001}. A few studies in the literature aim to evaluate the benefit of reducing the  uncertainty in the environment. In our prior work~\cite{yetkin.etal.oceans2015}, we show that inaccurate estimate of sensor performance can lead to inaccurate estimate of search performance. For example, when the presumed probability of detection is higher than the actual probability of detection, the probability that all objects have been found during a search mission is exaggerated, and the search mission might be terminated too early. For the particular trial in~~\cite{harris2005aqs}, the experimental results show that the mine-hunting mission takes $40\%$ less time when the environment is known compared to when there is no prior environmental information. 

We note that the existing literature on robotic exploration (see~\cite{Yamauchi.frontier.1997, bourgault.etal.exploration.2002, Makarenko.etal.exploration.2002, Carrillo.etal.exploration.2015}, among many examples) provides little insight to the applications we address. Robotic exploration addresses the challenge of building a map. In contrast, we seek to characterize a subset of the environment with respect to search sensor performance for the goal of improving search effectiveness. 

\section{Problem Formulation}
\label{sec:search}

Search and environmental characterization are accomplished using different sensors that can be mounted on different vehicles or on the same vehicle. When the sensors are placed on different vehicles, the vehicle that possesses the search sensor is called \emph{the search vehicle} and the vehicle that possesses the environmental characterization sensor is called \emph{the environmental characterization vehicle}. When the search and environmental characterization sensors operate simultaneously on a single vehicle, we informally refer to the vehicle as \emph{the search/environmental characterization vehicle}. In this section, we provide the notation and the formal definition of the search problem. A list of variables is provided in Table~\ref{tab:variables} in the Appendix. 

\subsection{Preliminaries}
We are given a search grid $\mathcal{G} \subset \mathbb{R}^2$ with $n_r$ rows and $n_c$ columns partitioned into $K = n_r n_c$ disjoint cells. We associate with each cell $i$ random variables $X_i$ and $E_i$ that represent the number of objects and the environmental conditions in the cell, respectively. We presume $X_i$ is independent of $X_j$ and $E_i$ is independent of $E_j$ when $i \neq j$. The objective of the search mission is to estimate $X_1, \ldots, X_K$ by using a sensor to detect objects in each cell. We assume that sensor performance is dependent on the environment, and we use a stochastic description of sensor performance in the environment. The environment in each cell is from a finite set of possible environments $w_1, w_2, \ldots, w_m$. In practice, clustering of different types of environmental conditions can be carried out by using a previously acquired environment dataset in the search domain (see, for example, \cite{McMahon2017}). We presume that the actual environmental condition in each cell is not known, but that a probability distribution is known for each cell. For instance, the environment probability distribution for $i$th cell is expressed $[p_1(i), p_2(i), \ldots, p_m(i)]$ where $p_j(i) = P(E_i = w_j)$ is the probability that the environment in cell $i$ is $w_j$. We note that sum of probabilities for each cell is unity,

\begin{equation}
\sum\limits_{j = 1}^m p_j(i) = 1
\end{equation}

\subsection{Sequential Bayesian update for the search vehicle}

When the search vehicle visits a cell, it acquires a noisy observation $z \in Z$ of the number of objects in the cell. We denote by $Z$ both the set of possible search measurements (\ie $z \in Z$) and the random variable associated with a search measurement in a cell (\ie $Z_i = z_i$). The observation $z$ may be less than the true number of objects because of missed detections, or it might be larger due to false alarms. In this study, we assume that the number of false alarms $z_f$ and correct detections $z_d$ are probabilistically independent. Hence, the value of the measurement $z$ can be expressed 
\[
	z = z_f + z_d
\]
We model the likelihood of observing $z$ objects when $x$ is the true number of objects given that the environment is $w_j$. The sensor model is 

\begin{align}
\P{ z \mid x, w_j} & = \hspace*{-2mm} \sum\limits_{l = 0}^{\text{min}(x,z)} \hspace*{-2mm} P_D \big ( z_d = l \mid x, w_j \big ) P_F \big ( z_f = z - l \mid w_j \big )
\label{eq:convolution}
\end{align}

\noindent where $P_D(z_d = l \mid x, w_j)$ is the probability that the sensor detects $l$ objects, and $P_F(z_f = k \mid w_j)$ is the probability that the sensor returns $k$ false alarms. The sensor model is also described in~\cite{yetkin.etal.oceans2015}, and is briefly presented here for clarity. Convolution of $P_D$ and $P_F$ in~\eqref{eq:convolution} follows from the assumption that the number of missed detections and false alarms are statistically independent. For numerical examples in Section~\ref{sec:numericalResults}, we model the probability of false alarms with a geometric distribution, and the probability of correct detections with a Binomial distribution,

\begin{align}
& P_F \big ( z_f = k \mid w_j \big ) =  (1 - F_j) F_j^k \quad \quad \quad k \geq 0 & \label{eq:falseAlarm} \\
& P_D \big ( z_d = l \mid x, w_j \big ) =  \binom {x} {l} D_j^l (1 - D_j)^{x-l} \hfill & 0 \leq l \leq x \nonumber
\end{align}

\noindent where $0 \leq F_j < 1$ denotes the probability of one or more false alarms, and $0 < D_j \leq 1$ denotes the probability of detection. Note that both $F_j$ and $D_j$ are assumed to vary as functions of the environment type $w_j$. Then, the likelihood is expressed

\begin{align}
\P{ z \mid x, w_j } & = \sum\limits_{k=0}^{\text{min}(x,z)} \binom {x} {k} D_j^k (1 - D_j)^{x-k} (1 - F_j ) F_j^{z-k}
\label{eq:likelihood_search}
\end{align}

In subsea applications, probability of false alarm and probability of detection are sometimes modeled through a receiver operating characteristics (ROC) curve which describes the probability of detection as a function of probability of false alarm \cite{baylog2015multiple, myers2010pomdp, zhang2001spectral}. We note that our intention in this study is not to model the characteristics of a specific sensor type. We believe the geometric distribution in~\eqref{eq:falseAlarm} efficiently models the intuition that fewer false alarms are more likely to occur than a greater number of false alarms. However, other expressions are also possible for the false alarm model, and our results do not depend on this specific false alarm model except for numerical illustrations.    

We assume the number of objects in each cell is statistically independent of the environmental conditions in that cell. Thus, $P(X = x \mid w_j) = P(X = x)$. We use Bayesian update law to update the distribution $P(X \mid z, w_j)$ when $z$ is observed.

\begin{equation}
\P{ X = x \mid z, w_j } = \frac{ \P{ z \mid x, w_j } \P{ X = x  }}{ \sum\limits_x \P{ z \mid X = x, w_j} \P{X = x} }
\label{eq:updated_search}
\end{equation}

\noindent where $P( X = x )$ is the prior distribution on the number of targets and $P(z \mid x, w_j)$ is the sensor characteristics as in~\eqref{eq:likelihood_search}. 

\subsection{Sequential Bayesian update for the environmental characterization vehicle}

When the characterization vehicle characterizes the environment at a location, it acquires the noisy observation $y \in Y$ of the true environment in the cell. We denote by $Y$ both the set of possible environment measurements (\ie $y \in Y$) and the random variable associated with an environment measurement at a cell (\ie $Y_i = y_i$). We assume the likelihood of observing a particular environment $y$ given the true environment $w_j$ is known before the characterization starts and it does not change. Insight on the form of the likelihood function arises from research on subsea bottom-type characterization, such as in~\cite{jaramillo2011auv}. We use a Bayesian update law to update the distribution of the environment when $y$ is observed,

\begin{equation}
\P{ E = w_j \mid y \in Y} = \frac{ \P{ y \mid E = w_j} \P{E = w_j} }{ \sum\limits_{j} \P{y \mid E = w_j} \P{E = w_j} }
\label{eq:updated_environment}
\end{equation}

\noindent where $P(E = w_j)$ is the prior probability that the environment at the location is $w_j$.

\subsection{Sequential Bayesian update for the search/environmental characterization vehicle}

When the search sensor and the environmental characterization sensor operate simultaneously on a single vehicle, the noisy observations $z$ and $y$ are acquired simultaneously. Given $z$ and $y$ measurements acquired at a location, we represent the updated probability distribution of the number of objects unconditioned on the environment

\begin{align}
\P{ X = x \mid z, y } = \sum\limits_{j} \P{ X = x \mid z, w_j } \P{ E = w_j \mid y }
\end{align}  

\noindent where the posterior distributions $P(X = x \ | \ z, w_j)$ and $P(E = w_j \ | \ y \in Y)$ follow from~\eqref{eq:updated_search} and~\eqref{eq:updated_environment}, respectively.

\section{Path Planning For The Search Vehicle}
\label{sec:searcher}

We perform environmental characterization to improve the results of a search mission. To better understand the value of acquiring an environment measurement at a location, we seek to quantify the effect of the acquired environment measurements on search results. Thus, in this section, we briefly present the value of searching a location and the objective function to compute the optimal search paths. For more details on path planning for the search vehicle, we refer the reader to our prior work \cite{yetkin.etal.oceans2015}.

The goal of a search mission is to maximize the probability that the estimated number of objects in a cell is correct. Thus we seek to maximize the estimation accuracy. After the search vehicle visits a location, we compute the estimate $\delta_X(z)$ of the number of objects $x$ at the location, based on the measurement $z$. When $\delta_X(z)$ is greater than $x$, we overestimate the number of objects, \ie we declare more than the actual number of objects are present. When $\delta_X(z)$ is less than $x$, we underestimate the number of objects, \ie we fail to declare some of the objects that are present. Both overestimation and underestimation may degrade the utility of the search results. Given the measured data $z$, we define the utility of the estimate $\delta_X(z)$ when $x$ is the true number of objects

\begin{equation}
\U{ x, \delta_X(z) } = \begin{cases} 1 & \mbox{if } x = \delta_X(z) \\
0 & \mbox{if } x \neq \delta_X(z) \end{cases}
\label{eq:utility_function}
\end{equation}

\noindent which penalizes the deviations from true number of objects. The zero-one function in~\eqref{eq:utility_function} emphasizes the fact that in some search missions, such as mine hunting, an incorrect estimate has no utility regardless of how close the estimate is. The posterior expected utility of computing the estimate $\delta_X(z)$ when the environment is $w_j$ is

\begin{equation}
\E{ \U{ x, \delta_X(z) } \mid z, w_j } = \sum\limits_{x} \P{ x \mid z, w_j} \U{x, \delta_X(z)}
\label{eq:expectedutility} 
\end{equation}

\noindent where the expectation is taken over the parameter space $X$ with respect to the posterior distribution $P(X = x \mid z, w_j)$. Let $\delta_X^\star(z)$ be the estimator that maximizes the expected utility in~\eqref{eq:expectedutility}. Such an estimator is called the Bayes estimator, and it is a function of the acquired measurement $z$ 

\begin{equation}
\delta_X^\star(z) = \arg\max_{\delta_X(z)} \ \E{ \U{ x, \delta_X(z) } \mid z, w_j }
\label{eq:bayesEstimator}
\end{equation}  

\noindent Then, the expected utility in~\eqref{eq:expectedutility} is the estimation accuracy conditioned on environment $w_j$ that we seek to maximize, when the estimator $\delta_X(z)$ is the Bayes estimator in~\eqref{eq:bayesEstimator}. Thus, the estimation accuracy conditioned on the environment $w_j$ after acquiring the measurement $z$ is 

\begin{equation}
\E{ \U{ x, \delta_X^\star(z) } \mid z, w_j } = \max_x \P{ X = x \mid z, w_j}
\label{eq:estimationaccuracy_z_wj}
\end{equation}

In order to assess the benefit of searching a cell, we compute estimation accuracy in~\eqref{eq:estimationaccuracy_z_wj} for each possible measurement $z \in Z$. This yields \emph{expected} estimation accuracy of searching a cell conditioned on environment $w_j$,  

\begin{equation}
\E{ \U{ x, \delta_X^\star(z) } \mid w_j } = \sum\limits_{z} \P{ z \mid w_j } \max_x \P{ x \mid z, w_j }
\label{eq:estimationaccuracy_wj}
\end{equation}

\noindent where $\P{z \mid w_j}$ is the probability of observing a particular measurement $z$ given the true environment $w_j$. Since deterministic knowledge on the environment is assumed to be unavailable, we compute the expected estimation accuracy unconditional on the environment. Averaging over the environments yields

\begin{equation}
\E{ \U{ x, \delta_X^\star(z) } } = \sum\limits_{j=1}^m \P{E = w_j } \E{ \U{ x, \delta_X^\star(z) } \mid w_j }
\label{eq:estimationaccuracy}
\end{equation}

\noindent and we call this the \emph{anticipated} estimation accuracy.  

Measurements from different cells are independent, and thus estimation accuracy for a path that passes through multiple cells is simply a product of the estimation accuracy for each cell in the path. Let $\gamma = \{q_1, \ldots, q_N \}$ be a candidate search path that traverses the cells $q_1, \ldots, q_N \in \mathcal{G}$, and let $\Cgamma$ be the budget constraint on the search mission due to limited time/distance the vehicle can traverse. We define $\mathcal{C}(\gamma)$ to denote the cost for traversing a path $\gamma$. Note that when the traversal cost for moving from one location to another is unity, $\mathcal{C}(\gamma)$ is simply the number of cells traversed by $\gamma$. 

When the vehicle makes multiple visits to a cell, we acquire a set of independent search measurements. We denote by $z$ both a single search measurement and a set of search measurements when a cell is visited multiple times by $\gamma$. Then, the expected utility of traversing $\gamma$ is

\begin{equation}
\begin{split}
\E{ \U{ x, \delta_X^\star(z_{\gamma}) } } = \prod_{q_i \in \gamma } \E{ \U{ & x_{q_i}, \delta_X^\star ( z_{q_i}) } } \\
& \times \prod_{i \in \mathcal{G} \setminus \gamma} \max_{x_i} \P{ X_i = x_i } 
\end{split}
\label{eq:pathutility}
\end{equation} 

\noindent where $\mathcal{G} \setminus \gamma$ denotes the remaining cells in the search grid that are not traversed by $\gamma$, and $\max_{x_i} \P{ X_i = x_i }$ is the certainty in the number of objects in cell $i$ prior to acquiring new measurements. Let $\Omega_\gamma$ denote the finite collection of feasible search paths. Then, the optimal search path is

\begin{equation}
\gamma^\star = \arg\max_{\gamma \in \Omega_\gamma} \E{ \U{ x, \delta_X^\star(z_{\gamma}) } }
\label{eq:bestPath_searchonly}
\end{equation}

\noindent subject to

\begin{equation}
\mathcal{C}(\gamma) \leq \Cgamma
\end{equation}

\section{Path Planning For The Environmental Characterization Vehicle}
\label{sec:characterization}

The primary objective of environmental characterization is to improve search performance. With additional information about the environment at a few locations, it might be possible to avoid searching locations where the sensor performs poorly in favor of places where the sensor performs well. We consider the case where environmental characterization is performed prior to search, and we assume that both environmental characterization and search cannot be performed exhaustively due to limited resources. 




\subsection{Entropy change maximization}
\label{sec:entropy}

When environment information can be acquired only in some locations due to limited resources, the question is to determine where to optimally sample the environment. One approach that is often used in similar applications is to maximize the change in entropy due to acquired environment measurements~\cite{papadimitriou2000entropy, coleman2007nonlinear, elfes1992dynamic}. We briefly describe a typical entropy approach for selecting where to sample the environment so that we can compare it to our proposed approach.

Let $H(E = w_j)$ denote the prior entropy of the probability distribution $P(E = w_j)$ and let $H(E = w_j \mid y \in Y)$ be the posterior entropy after acquiring the environment measurement $y$. The expected amount of change in the entropy for a future environment measurement $y$ can be computed by


\begin{align*}
J \big ( E \big ) = \H{ E = w_j} - \sum\limits_y \H{ E = w_j \mid y \in Y}
\end{align*}

Let $\eta$ be a candidate path for the environment characterization vehicle, $\Omega_\eta$ be the finite collection of feasible paths, and $\Ceta$ be the budget constraint on the environment characterization mission. Then, the best path to characterize the environment based on the entropy change maximization method is 

\begin{align}
\eta^\star = \arg\max_{\eta \in \Omega_\eta} J_{\eta}(E)
\label{eq:characterizationpath_entropy}
\end{align}

\noindent subject to

\begin{equation*}
\mathcal{C}(\eta) \leq \Ceta
\end{equation*}

However, we note that the purpose of environmental characterization in this study is not to explore the environment, but to improve the performance of a follow-on search mission. We show in Section~\ref{sec:numericalResults} via numerical studies that maximizing change in entropy does not maximize the performance of follow-on search missions. 

\subsection{Environmental loss function}
\label{sec:lossFunc}

Due to the uncertainty in the environment and the noise in environmental observations, estimation accuracy after visiting a cell in~\eqref{eq:estimationaccuracy} may be different than actual estimation accuracy if the true environment were unambiguously known. In our prior work \cite{yetkin.etal.oceans2015}, we discuss the effect of environment uncertainty on search results, and show that deviations from the true environment result in deviations from actual estimation accuracy and degrade the search performance. In this paper, we extend our findings in~\cite{yetkin.etal.oceans2015} to select the best locations to conduct environmental surveys. Our approach is to define a linear loss function that penalizes deviations from the actual expected estimation accuracy for each cell. 

To formally define the loss function, we first introduce a preference ordering $\preceq$ on environments. Suppose there is a finite set of environments $\{ w_1, \ldots, w_m \}$. For the notational convenience, let $\V{w_j}$ denote the expected estimation accuracy conditioned on environment $w_j$    

\begin{equation}
\V{w_j} = \E{ \U{ x, \delta_X^\star(z) } \mid E = w_j }
\label{eq:V_EEA}
\end{equation}

\noindent in~\eqref{eq:estimationaccuracy_wj}. We say the environment $w_i$ is more preferred for the search than the environment $w_j$ if the expected estimation accuracy conditioned on $w_i$ is greater than the expected estimation accuracy conditioned on $w_j$. That is, we say that $w_i \preceq w_j$ if and only if $V(w_i) \leq V(w_j)$. If for some $w_i, w_j$ when $i \neq j$ we have $V(w_i) = V(w_j)$, then $w_i = w_j$. If $V(w_i) \neq V(w_j)$, we say $w_i$ and $w_j$ are distinct environments. Suppose the environments $w_1, \ldots, w_m$ are distinct and \emph{ordered} so that $w_1 \prec w_2 \prec \ldots \prec w_m$, let $e$ be the true environment in a cell, and let $\delta_E(y)$ be an estimate of the environment based on measurement $y$. When the true environment is $e$, the loss due to the estimate $\delta_E(y)$ is defined

\begin{equation}
L_V\loss{e,\delta_E(y)}  =
 \begin{cases}
   c_1 \Big (  \V{ e } - \V{ \delta_E(y) } \Big ) &  \text{if} \ \delta_E(y) \preceq e \vspace*{3mm}\\
   c_2 \Big (  \V{ \delta_E(y) } - \V{ e } \Big ) &  \text{if} \ \delta_E(y) \succ e
 \end{cases}
 \label{eq:lossFunction}
\end{equation}    

\noindent where $c_1, c_2 > 0$ are the relative costs of over and underestimation. Underestimating the environment, $\delta_E(y) \prec e$, may result in unnecessary extra visits to improve the belief on the number of objects at a location. However, overestimating the environment, $\delta_E(y) \succ e$, may yield to inaccurate estimates on the number of objects. In some search applications, such as mine-hunting, overestimation is less preferred to underestimation. Thus, we may assign the relative costs such that $c_1 < c_2$.       

Given the environment measurement $y$, the posterior expected loss of computing the environment estimate $\delta_E(y)$ is

\begin{align}
\E{ L_V\loss{ e, \delta_E(y) } \mid Y = y } = \sum\limits_{j = 1}^m \P{ w_j \mid y } L_V\loss{ w_j, \delta_E(y) }
\label{eq:expectedLoss}
\end{align}

\noindent where $\P{w_j | y}$ is the updated probability that $w_j$ is the environment at the location after observing the environmental measurement $y$. We choose the estimator that minimizes the expected loss in~\eqref{eq:expectedLoss}. Let $\delta_E^\star(y)$ be the Bayes estimator such that 

\begin{align}
\delta_E^\star(y) = \arg\min_{\delta_E(y)} \E{ L_V\loss{ e, \delta_E(y) } \mid y }
\label{eq:bayesEstimatorForLoss}
\end{align}

For the specific loss function in~\eqref{eq:lossFunction}, the Bayes estimator that minimizes the expected loss can be computed from the cumulative distribution (see~\cite{berger2013statistical} for more details), 

\begin{eqnarray}
 F_{E\mid y}(w_n) &:= & \P{ E\preceq w_n \mid Y = y } \nonumber \\
 & = &  \sum_{ i=1 }^{n}  \P{ E = w_i \mid Y = y } \label{eq:cumulativeIsSum}
\end{eqnarray}
for $n\in\{1,2,\dots,m\}$ where~\eqref{eq:cumulativeIsSum} follows since $w_1 \prec w_2 \prec \cdots \prec w_m$. For
\begin{equation*}
 \ell = \max \left\{ n\in\{1,2,\dots,m-1\} : F_{E\mid y}(w_n) \leq \frac{c_1}{c_1+c_2} \right\} \ , \label{eq:ellDef}
\end{equation*}

\noindent the optimal estimate is
\begin{equation}
 \delta_E^\star(y) =  w_{\ell+1}
 \label{eq:optimalEhat}
\end{equation}

\subsection{Path planning}
\label{sec:pathPlanning}

A benefit of environmental surveys is to reduce the error in anticipated search performance due to uncertainty in the environment. When a location is not visited by the search vehicle during a search mission, acquiring an environment measurement at that location will not affect search performance. From the loss function in~\eqref{eq:lossFunction}, computing the estimate~\eqref{eq:bayesEstimatorForLoss} of the environment after acquiring environment measurement $y$ yields the conditional expected loss

\begin{equation}
\E{ L_V\loss{ e, \delta_E^\star(y) } \mid y } = \sum\limits_{j = 1}^m \P{ E = w_j \mid y } \ L_V\loss{ w_j, \delta_E^\star(y) }
\label{eq:condExpLoss}
\end{equation}

\noindent that quantifies the amount of uncertainty in anticipated estimation accuracy after acquiring $y$. Informally speaking, the prior loss before acquiring an environment measurement represents the prior uncertainty, and the conditional expected loss in~\eqref{eq:condExpLoss} represents the posterior uncertainty in search performance. For notational convenience, we define $\mathcal{R}(y)$ to denote the reduction of uncertainty in anticipated estimation accuracy due to environment measurement $y$
\begin{equation*}
\mathcal{R}(y) = \E{ L_V\loss{ e, \delta_E^\star } } - \E{ L_V\loss{ e, \delta_E^\star(y) } \mid y \in Y }
\end{equation*} 

\noindent Then, the gain of acquiring an environment measurement $y_i$ in cell $i$ is the reduction of uncertainty in anticipated estimation accuracy given that cell $i$ is visited by the search vehicle

\begin{align}
\begin{split}
G \big ( y_i \big ) = \I{i, \gamma^\star(y_i)} \ \mathcal{R}(y_i) 
\end{split}
\label{eq:U2(y)}
\end{align}

\noindent where the notation $\gamma^\star(y)$ denote the best path for the search vehicle when the probability distribution of the environment is updated with the acquired measurement $y$, and the indicator function $\I{i, \gamma(y_i)}: y_i \rightarrow [ c^\prime, 1]$ is defined

\begin{equation}
\I{i, \gamma(y_i)}  =
 \begin{cases}
   1 &  i \in \gamma(y_i) \\
    c^\prime  &  i \not \in \gamma(y_i)
 \end{cases}
 \label{eq:indicator}
\end{equation}

\noindent where $0 \leq c^\prime < 1 $ is a parameter to determine the relative gain of sampling the environment at locations that will not be searched. Without loss of generality, we consider that $c^\prime = 0$.

Let $\eta = \{q_1, q_2, \ldots, q_M \}$ be a candidate path for the environment characterization vehicle and recall that $\Omega_\eta$ is the finite collection of feasible characterization paths. We denote by $y$ both a single environment measurement and a set of independent environment measurements when a cell is visited multiple times by $\eta$. Then, the expected characterization gain of traversing $\eta$ is

\begin{align}
\E{ G_\eta } = & \sum\limits_{y_{\eta} } \P{ Y_{\eta} = y_{\eta} } \sum\limits_{q_i \in \eta }  \I{i, \gamma^\star(y_i)} \ \mathcal{R}(y_{q_i}) 
\label{eq:pathGain}
\end{align}

\noindent and the optimal path is 

\begin{equation}
{\eta}^\star = \arg\max_{\eta \in \Omega_\eta } \E{ G_\eta }
\label{eq:bestPath_searchonlyAfterY}
\end{equation}

\noindent subject to

\begin{equation}
\mathcal{C}(\eta) \leq \Ceta
\label{eq:budgetConstraint_eta}
\end{equation}

\subsection{Approximating the characterization gain of a path}
\label{sec:approximation_path}

Computing the optimal path for the environment characterization vehicle in~\eqref{eq:bestPath_searchonlyAfterY} can be computationally very expensive. This is mainly due to large computational requirements of computing the optimal search path for each set of environment measurements along a candidate path $\eta$. Thus, we also propose an approximate method that reduces the computational complexity of the solution in~\eqref{eq:bestPath_searchonlyAfterY}. Our approximate solution yields provably near-optimal paths.

Let $\bar{\Omega}_\eta$ be the set of environment characterization paths such that $\mathcal{C}(\eta) \leq \Ceta$, let $ \lvert \cdot \rvert$ denote the size of a set (or an array), and let $\mathcal{S}$ denote the computational complexity of computing the optimal search paths. Then, the solution in~\eqref{eq:bestPath_searchonlyAfterY} has a computational complexity of $\mathcal{O} \big ( \lvert \bar{\Omega}_\eta \rvert m^{\lvert \eta \rvert} \mathcal{S} \big ) $ where $m$ is the number of environments in the search domain. This shows that the exponential increase in the computational complexity is dominated by the large planning horizon for the characterization vehicle. One approach to reduce this computational complexity is to use a receding horizon strategy where we compute the paths for a shorter horizon. While receding horizon approach may require less computational power compared to computing the paths for the entire planning horizon, it may still be infeasible unless the considered planning horizon is sufficiently small (in which case the resulting performance will be poor). Instead, our approach to reduce the complexity of the solution in~\eqref{eq:bestPath_searchonlyAfterY} aims to approximate the characterization gain of traversing a path.

We start with re-arranging the terms in~\eqref{eq:pathGain} by partitioning a path $\eta$ into two parts; a cell $q_i \in \eta$ and the other cells in the path

\begin{align}
\Es{ G_\eta } & = \sum\limits_{y_{\eta} } \P{ Y_{\eta} = y_{\eta} } \sum\limits_{q_i \in \eta }  \I{q_i, \gamma^\star(y_\eta)} \ \mathcal{R}(y_{q_i}) \nonumber \\
& = \sum\limits_{q_i \in \eta } \sum\limits_{y_{q_i} } \ \mathcal{R}(y_{q_i}) \sum\limits_{y_{\eta \setminus q_i} } \I{q_i, \gamma^\star(y_\eta)} \ \P{ Y_{\eta} = y_{\eta} }
\label{eq:pathGain_arranged}
\end{align}

\noindent where $\eta \setminus q_i$ denotes the set of cells in $\eta$ except cell $q_i$. The terms up to the third summation in~\eqref{eq:pathGain_arranged} denote the characterization gain of acquiring the environment measurement $y$ from cell $q_i \in \eta$, and the other terms that start with the third summation denote how likely it is that sampling cell $q_i$ will improve the performance of a follow-on search mission. That is, it represents the chances that cell $q_i$ will be visited during a follow-on search mission based on the environment measurements that we may acquire along the path $\eta$. Note that since $\eta = \{q_1, q_2, \ldots, q_M \}$, the joint probability $\P{ Y_{\eta} = y_{\eta} }$ in~\eqref{eq:pathGain_arranged} can be expressed

\begin{equation}
\P{ Y_{\eta} = y_{\eta} } = \P{ Y_{q_1} = y_{q_1} } \times \cdots \times \P{ Y_{q_M} = y_{q_M} }
\end{equation}

Thus, we can re-write~\eqref{eq:pathGain_arranged}

\begin{align}
\Es{ G_\eta } = \sum\limits_{q_i \in \eta } \sum\limits_{y_{q_i} } \Big ( \P{ Y_{q_i} = y_{q_i} } \mathcal{R}(y_{q_i}) \Big ) \times \bar{\mathbf{P}}_{\eta \setminus q_i}
\label{eq:pathGain_arranged2}
\end{align}

\noindent where $\bar{\mathbf{P}}_{\eta \setminus q_i}$ is the total probability of every possible sets of the environment measurements $y_{q_1}, y_{q_2}, \ldots, y_{q_{i-1}}, y_{q_{i+1}}, \ldots, y_{q_M}$ such that the optimal search path associated with the updated environment distributions visits cell $q_i$

\begin{align}
\bar{\mathbf{P}}_{\eta \setminus q_i}  & = \sum\limits_{y_{\eta \setminus q_i} } \I{q_i, \gamma^\star(y_\eta)} \ \P{ Y_{\eta \setminus q_i} = y_{\eta \setminus q_i} } \\
& = \sum\limits_{ y_{\eta \setminus q_i} \mathbf{:} \ q_i \in \gamma^\star (y_\eta)} \P{ Y_{\eta \setminus q_i} = y_{\eta \setminus q_i} }
\label{eq:Pbar}
\end{align}

Indeed, the computational cost of~\eqref{eq:pathGain_arranged2} is dominated by $\bar{\mathbf{P}}_{\eta \setminus q_i}$ in~\eqref{eq:Pbar}. Thus, we use a sample-based method to compute an empirical estimate of $\bar{\mathbf{P}}_{\eta \setminus q_i}$, which results in a significant speed-up in computing the characterization path. For each cell $q_i \in \eta$ and for every environment measurement $y_{q_i} \in Y$, we perform $\bar{N}$ trials where, in each trial, we randomly sample an environment measurement from the probability distribution $\P{Y_{q_j} = y_{q_j}}$ for all $q_j \in \eta$ such that $j \neq i$. Then, with the updated environment distributions we compute the optimal search path and see whether the corresponding path visits cell $q_i$, or not. We simply count the number of times cell $q_i$ is being visited by the resulting search path out of $\bar{N}$ trials, and we denote this number by $k$. Since this is repeated for every other cell in the path, we may sample the same environment measurement $y_{q_i}$ from cell $q_i$ during a trial of another cell. Let $\bar{N}_{y_{q_i}}$ be the number of times $y_{q_i}$ is sampled in cell $q_i$ during the trials of the other cells in path $\eta$ and let $k_{y_{q_i}}$ be the number of times cell $q_i$ is contained in the corresponding search path out of these $\bar{N}_{y_{q_i}}$ trials. Then, the empirical estimate of $\bar{\mathbf{P}}_{\eta \setminus q_i}$ is

\begin{equation}
\hat{\mathbf{P}}_{\eta \setminus q_i} = \frac{k + k_{y_{q_i}}}{\bar{N} + \bar{N}_{y_{q_i}}}
\end{equation}

Obtaining a close estimate of $\bar{\mathbf{P}}_{\eta \setminus q_i}$ is important to compute a near-optimal characterization path. We show that a bound on the distance between $\bar{\mathbf{P}}_{\eta \setminus q_i}$ and $\hat{\mathbf{P}}_{\eta \setminus q_i}$ can be computed. We first note that after each trial for a cell, that cell is either contained in the follow-on search path, or it is not contained. Thus, we can cast each trial as a Bernoulli trial where the result of the trial is either 1 if the cell is contained in the search path, or it is 0 if the cell is not contained. Then, we use Hoeffding's inequality~\cite{hoeffding1963probability} to obtain a probabilistic bound on the difference between $\bar{\mathbf{P}}_{\eta \setminus q_i}$ and $\hat{\mathbf{P}}_{\eta \setminus q_i}$

\begin{equation}
\P{ | \bar{\mathbf{P}}_{\eta \setminus q_i} - \hat{\mathbf{P}}_{\eta \setminus q_i} | < \epsilon \bar{\mathbf{N}} } \geq 1 - 2 \exp ^{-2\epsilon^2 \bar{\mathbf{N}} }
\label{eq:hoelfding_Pbar}
\end{equation}

\noindent where $\bar{\mathbf{N}} = \bar{N} + \bar{N}_{y_{q_i}}$ and $\epsilon > 0$. Replacing  $\bar{\mathbf{P}}_{\eta \setminus q_i}$ with its estimate $\hat{\mathbf{P}}_{\eta \setminus q_i}$ in~\eqref{eq:pathGain_arranged2} approximates the characterization gain of a path. We denote the approximate characterization gain of a path $\eta$ by $\Es{\hat{G}_\eta}$ 

\begin{align}
\Es{ \hat{G}_\eta } = \sum\limits_{q_i \in \eta } \sum\limits_{y_{q_i} } \frac{k + k_{y_{q_i}}}{\bar{N} + \bar{N}_{y_{q_i}}} \P{ Y_{q_i} = y_{q_i} } \mathcal{R}(y_{q_i}) 
\label{eq:pathGain_approximate}
\end{align}

\noindent Finally, we select the path that maximizes the approximate characterization gain in~\eqref{eq:pathGain_approximate} subject to the budget constraint in~\eqref{eq:budgetConstraint_eta}.

\begin{equation}
\hat{\eta} = \arg\max_{\eta \in \Omega_\eta } \Es{ \hat{G}_\eta}
\label{eq:approximate_optimalpath}
\end{equation}

Now, we define the following theorem and the corollary, where we first bound the difference between the characterization gain and the approximate characterization gain for a path, and we then bound the difference between the optimal characterization gain and the approximately optimal characterization gain. Proofs for both the theorem and the corollary are provided in the Appendix. 

\begin{theorem}
The probability of the difference between the characterization gain~\eqref{eq:pathGain_arranged2} and its estimate~\eqref{eq:pathGain_approximate} for a path $\eta$ satisfying a specific bound is expressed   
\begin{equation}
\P{ \big | \Es{G_\eta} - \Es{\hat{G}_\eta} \big | < \epsilon \bar{\mathbf{N}} |\eta| } \geq 1 - 2 \exp ^{-2\epsilon^2 \bar{\mathbf{N}} }
\label{eq:hoelfding_characterizationgain}
\end{equation}
\label{theorem:bound_on_gain}
\end{theorem}

\begin{corollary}
For the optimal characterization path $\eta^\star$ in~\eqref{eq:bestPath_searchonlyAfterY} and the approximate path $\hat{\eta}$ in~\eqref{eq:approximate_optimalpath}, the difference in expected characterization gain satisfies
\begin{equation}
\P{ \Es{G_{\eta^\star}} - \Es{G_{\hat{\eta}}} < 2 \epsilon \bar{\mathbf{N}} |\eta| } \geq 1 - 2 \exp ^{-2\epsilon^2 \bar{\mathbf{N}} }
\label{eq:hoelfding_characterizationgain}
\end{equation}
\label{theorem:bound_on_optimalgain}
\end{corollary}

The proposed approximation approach yields a provably near-optimal path for the environment characterization vehicle. The computational complexity of the solution is reduced from $\mathcal{O} \big ( \lvert \bar{\Omega}_\eta \lvert m^{\lvert \eta \lvert} \mathcal{S} \big ) $ to $\mathcal{O} \big ( \lvert \bar{\Omega}_\eta \lvert { \lvert \eta \lvert} m \bar{N} \mathcal{S} \big ) $. In general, choosing a larger value for $\bar{N}$ is likely to reduce the approximation error. However, our preliminary results show that a small value for $\bar{N}$ is often sufficient to obtain a close approximation. 

\subsection{Approximating the characterization gain of a cell}
\label{sec:approximation_cell}

In this section, we introduce an alternative approach to approximate the optimal characterization gain. Our alternative approach assumes that the search paths are composed of sequences of parallel straight lines. This assumption arises often in subsea applications that rely on side-scan imaging sonar (see, for example,~\cite{Houston2008,Hayes2009}). Unlike the proposed approach in Section~\ref{sec:approximation_path}, our approach in this section can only apply to certain classes of mapping problems.

Suppose the search area $\mathcal{G}$ consists of a set of parallel straight lines (a line is either a row or a column) $l_1, l_2, \ldots, l_{n_l}$. When each line corresponds to a row, $n_l = n_r$, and when each line corresponds to a column, $n_l = n_c$. Suppose that the $j$th line traverses the cells $q_{j1}, q_{j2}, \ldots, q_{jk}$. Thus, when the vehicle traverses line $l_j$, it samples from cells $q_{j1}, q_{j2}, \ldots, q_{jk}$. Let $\eta$ be a candidate path for the environment characterization vehicle that consists of lines $l_1, l_2, \dots l_k$, and consider that cell $i \in \eta$ is contained in $l_j$. We claim that the value of characterizing cell $i$ along path $\eta$ can be closely approximated by the value of characterizing cell $i$ along line $l_j$. That is

\begin{align}
\E{ G_{q_i \in \eta} } & = \sum\limits_{y_{q_i} } \Big ( \P{ Y_{q_i} = y_{q_i} } \mathcal{R}(y_{q_i}) \Big ) \times \bar{\mathbf{P}}_{\eta \setminus q_i} \label{eq:gain_path_cell} \\
& \approx \sum\limits_{y_{q_i} } \Big ( \P{ Y_{q_i} = y_{q_i} } \mathcal{R}(y_{q_i}) \Big ) \times \bar{\mathbf{P}}_{l_j \setminus q_i} \label{eq:gain_line_cell}
\end{align}

\noindent where

\begin{equation}
\bar{\mathbf{P}}_{l_j \setminus q_i}  = \sum\limits_{ y_{l_j \setminus q_i} \mathbf{:} \ q_i \in \gamma^\star (y_{l_j})} \P{ Y_{l_j \setminus q_i} = y_{l_j \setminus q_i} }
\label{eq:Pbar_line}
\end{equation}

Due to~\eqref{eq:gain_line_cell}, we can compute the value of characterizing a particular cell by only looking at the cells in the associated line (a row or a column). By doing so, we can compute the characterization gain for each cell individually. Then, the characterization gain of a path is simply the summation of the characterization gain of each cell in that path. We note that we do not have a formal guarantee of how closely~\eqref{eq:gain_line_cell} approximates~\eqref{eq:gain_path_cell}. However, our initial tests as well as the intuition suggest that~\eqref{eq:gain_path_cell} can be well-approximated by~\eqref{eq:gain_line_cell}.     

Since computing $\bar{\mathbf{P}}_{l_j \setminus q_i}$ in~\eqref{eq:Pbar_line} can be very expensive when the length of a line is large, we instead compute an empirical estimate of $\bar{\mathbf{P}}_{l_j \setminus q_i}$ as described in Section~\ref{sec:approximation_path}. For each cell in the search grid, we perform $\bar{N}$ trials to compute an empirical estimate of $\bar{\mathbf{P}}_{l_j \setminus q_i}$, where in each trial we sample environment measurements from the remaining cells in line $l_j$. After approximating the characterization gain for each cell individually, we apply an exact branch-and-bound method (similar to our prior work in~\cite{McMahon2017}) to compute the near-optimal path for the characterization vehicle. Suppose that the complexity of computing the characterization path when cell-wise characterization gains are known is similar to that of the search path. Then, this approximation approach yields a $\mathcal{O} \big ( (r m \bar{N} + 1) \mathcal{S} \big ) $ complexity of computing the near-optimal characterization paths.   

\section{Path Planning For The Search/environmental characterization Vehicle}
\label{sec:compound}

We lastly consider the case that a single vehicle is equipped with an environmental characterization sensor and a search sensor, and that both sensors can operate simultaneously. We again seek to maximize estimation accuracy. Unlike Section~\ref{sec:searcher} where the search vehicle aims to maximize the estimation accuracy with only the search measurements, we now acquire both a search measurement $z$ and an environmental measurement $y$ when the vehicle visits a location. Thus, the path strategy in Section~\ref{sec:searcher} that do not address the acquisition of environmental measurements do not apply to this case. 

Let $\W{w_j}$ denote estimation accuracy conditioned on the environment $w_j$ in~\eqref{eq:estimationaccuracy_wj},   

\begin{equation}
\W{w_j} = \max_x \P{ X = x \mid z, w_j }
\label{eq:W_EA}
\end{equation} 

\noindent and let $\{w_1, w_2, \ldots, w_m \}$ be a set of environments. We say $w_i \preceq w_j$ if and only if $\W{w_i} \leq \W{w_j}$. Note $\W{w_j}$ in~\eqref{eq:W_EA} is the accuracy of the estimate of the number of objects at a location while $\V{w_j}$ in~\eqref{eq:V_EEA} is the expected accuracy when a measurement $z$ has not yet been acquired. Suppose the environments $w_1, \ldots, w_m$ are distinct and \emph{ordered} (as defined in Section~\ref{sec:lossFunc}) so that $w_1 \prec w_2 \prec \ldots \prec w_m$, let $e$ be the true environment in a cell, and let $\delta_E(y)$ be an estimate of the environment based on the environment measurement $y$. When the true environment is $e$, the loss due to the estimate $\delta_E(y)$ is defined

\begin{equation}
L_W\loss{ e, \delta_E(y) }  =
 \begin{cases}
    c_1 \big ( \W{ \delta_E(y) } - \W{ e } \big ) &  \text{if} \ \delta_E(y) \succ e \\
    c_2 \big ( \W{ e } - \W{ \delta_E(y) } \big)  &  \text{if} \ \delta_E(y) \preceq e
 \end{cases}
 \label{eq:lossFunctionCompound}
\end{equation}    

\noindent where $c_1, c_2 > 0$ are again the relative costs of over and underestimation. Then, the posterior expected loss of computing the environment estimate $\delta_E(y)$, and the corresponding Bayes estimator $\delta^\star_E(y)$ are

\begin{align}
\E{ L_W\loss{ e, \delta_E(y) } \mid z, y } = \sum\limits_{j = 1}^m \P{ w_j \mid y } L_W\loss{ w_j, \delta_E(y) }
\label{eq:expectedLossCompound}
\end{align}

\begin{align}
\delta^\star_E(y) = \arg\min_{\delta_E(y)} \E{ L_W\loss{ e, \delta_E(y) } \mid z, y }
\label{eq:bayesEstimatorForLossCompound}
\end{align}

Given measurement $z$ and the estimate $\delta^\star_E(y) \in \{w_1, \ldots, w_m \}$ from  measurement $y$, the probability that the estimate of the number of objects at a location is correct is computed from
\begin{equation}
\E{ \U{ x, \delta_X(z) } \mid z, \delta^\star_E(y) } = \max_x \P{ X = x \mid z, \delta^\star_E(y) }
\label{eq:estimateAccuracyEstimate}
\end{equation}

\noindent In order to assess the benefit of visiting a location, we compute the \emph{estimated} estimation accuracy in~\eqref{eq:estimateAccuracyEstimate} for each possible set of observations $z \in Z, \ y \in Y$. Then, the expected estimation accuracy before visiting a location can be computed 

\begin{equation}
\E{ \W{  \delta_E(y) } } = \sum\limits_z \sum\limits_y \P{ z, y } \ \max_x \P{ X = x \mid z, \delta^\star_E(y) }
\label{eq:EEAgZY}
\end{equation}

\noindent where

\begin{align}
\P{ z, y} = \sum\limits_x \sum\limits_{w_j} \P{ z \mid x, w_j} \P{y \mid w_j} \P{x} \P{w_j} 
\label{eq:PzyJoint}
\end{align}

We again consider the candidate search path $\gamma$, the finite collection of feasible search paths $\Omega_\gamma$, and the budget constraint $\Cgamma$ on the vehicle. Let $y_{q_i}$ be the set of independent environment measurements acquired at $q_i$th cell. The expected estimation accuracy for traversing $\gamma$ is 

\begin{align}
\E{ \W{ \delta_E(y_\gamma) } } = \hspace*{-0.1cm} \prod_{q_i \in \gamma } \E{ \W{ \delta_E(y_\gamma) } } \times \hspace*{-0.1cm} \prod_{i \in \mathit{S} \setminus \gamma} \max_{x_i} \P{ x_i } 
\label{eq:expUtilPathCompound}
\end{align} 

\noindent and the optimal path is

\begin{equation}
\gamma^\star = \arg\max_{\gamma \in \Omega_\gamma} \E{ \W{ \delta_E(y_\gamma) } }
\label{eq:bestPath_searchonlyCompound}
\end{equation}

\noindent subject to

\begin{equation}
\mathcal{C}(\gamma) \leq \Cgamma
\end{equation}

%
%
%
%
%
%
%
%
%

\section{Numerical Results}
\label{sec:numericalResults}

In this section, we present simulation results that show the efficacy of the proposed search and environmental characterization strategies. Our numerical illustrations aim to evaluate search performance when environmental measurements are available for simplistic scenarios that are inspired by subsea mine-hunting missions. We present numerical illustrations for two scenarios. In one case, search and environmental characterization sensors are on different vehicles and environmental characterization is performed prior to search.  In the other case, search and environmental characterization sensors are on the same vehicle and both activities occur simultaneously. When each sensor operate on separate vehicles, our proposed approach maximizes the reduction of uncertainty in search performance~\eqref{eq:pathGain}. Thus, our approach should, on average, display less anticipated estimation accuracy error than other approaches. 

We divide the bounded search area $\mathcal{G}$ into a grid with $10\times 10$ non-intersecting cells. For each cell $i \in \mathcal{G}$, we assume there is $0 \leq x_i \leq L$ number of objects bounded above by $L$. In mine hunting missions, $x_i$ represents the number of mine-like objects residing in cell $i$. It is assumed that no prior information exists about the number of objects in any cell. We note that $L$ is typically not known beforehand; however, letting $L$ be a sufficiently large number will capture all realistic scenarios. In our simulations, $L = 2$. The performance of the search sensor is dependent on the environmental conditions. The particular sensor model that we use for the numerical illustrations is~\eqref{eq:likelihood_search}. We assume there are three candidate environments in the search area, $w_1, w_2, w_3$. The probability of detection, $D$, and the probability of at least one false alarm, $\alpha$, for each environment are $D = 0.65$ and $\alpha = 0.4$ for environment $w_1$, $D = 0.8$ and $\alpha = 0.3$ for environment $w_2$, and $D = 0.95$ and $\alpha = 0.05$ for environment $w_3$. Note that the information about the number of objects revealed after searching a cell increases with increasing probability of detection and decreases with increasing probability of false alarm. Thus, environment $w_1$ is the least and environment $w_3$ is the most informative. We consider that the sensor model for environment characterization is 
\begin{equation}
a_{ij} = \P{Y = w_i \mid E = w_j} \ \  \text{for all } \ i,j \in \{1, 2, 3 \} 
\label{eq:sensormodel_environment}
\end{equation}
\noindent where $a_{ii}$ is the probability of observing the true environment $w_i$. For the numerical illustrations, we use the characterization sensor model with $a_{11} = 0.9$, $a_{22} = 0.92$, $a_{33} = 0.94$. That is, for example, there is 0.9 probability of acquiring environment measurement $w_1$ when $w_1$ is the true environment at the location. The noisy environment observations are due to nonzero probabilities of observing environment $w_i$ when true environment is $w_j$, denoted by $a_{ij}$ for $i \neq j$. We assume the probability of acquiring incorrect environment measurement is the same for all possible environments other than the true environment. For example, when the true environment at a location is $w_1$, since $a_{11} = 0.9$, the probability of acquiring environment measurement $w_2$ and probability of acquiring environment measurement $w_3$ are $a_{21} = a_{31} = 0.05$. 

In the subsea applications that inspire our numerical illustrations, autonomous underwater vehicles (AUVs) are typically equipped with a side scan-sonar. Because side-scan sonar works poorly while the vehicle is turning, we associate a cost with vehicle turns and constrain the motion of the vehicle in a way that the vehicle can only move forward towards the next grid cell in the row. In order to account for the effect of turns, the vehicle passes through cells that are outside of the search area when transitioning between rows. Passing through cells that are outside the search area requires time but does not improve estimation accuracy since no measurements are acquired. For the numerical illustrations, we consider a unit cost for moving from a cell to an adjacent cell. Thus, the total cost of a mission is the length of the planning horizon, which we refer to as the mission length.  

\subsection{Numerical illustrations}
\label{sec:illustration1}

\begin{figure}[t!]
\hspace*{-0.3cm}
\begin{minipage}{\textwidth}
\vspace*{0.5cm}
\begin{minipage}[b]{.25\textwidth}
\includegraphics[width=0.85\textwidth,keepaspectratio,clip]{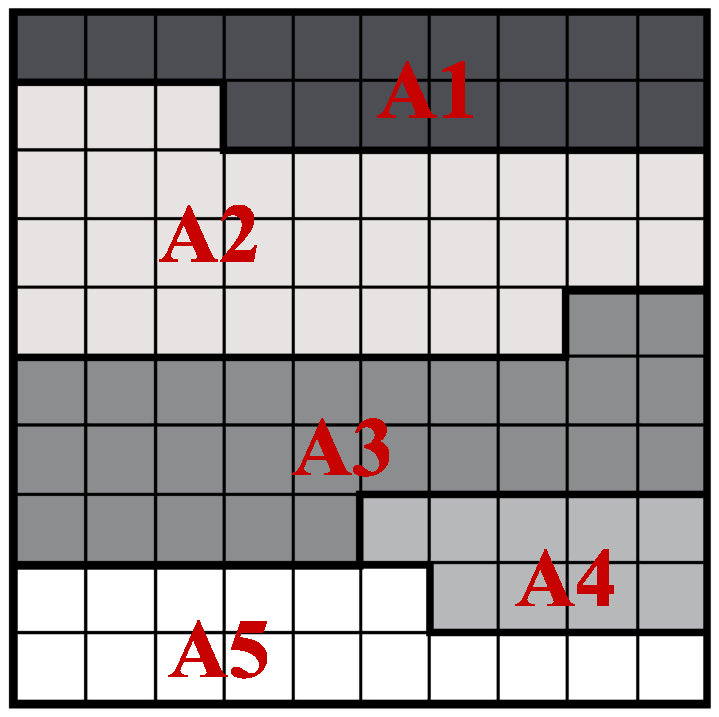}
\end{minipage}
\begin{minipage}[b]{.2\textwidth}
    \begin{tabular}{l c l }\hspace*{-0.3cm} \vspace*{-3.4cm}
\mbox{} & \mbox{} \\ \hline 
A1 & $\rightarrow$ & $[1.00, 0.00, 0.00]$\\
A2 & $\rightarrow$ & $[0.15, 0.20, 0.65]$\\
A3 & $\rightarrow$ & $[0.40, 0.30, 0.30]$\\
A4 & $\rightarrow$ & $[0.15, 0.80, 0.05]$\\
A5 & $\rightarrow$ & $[0.05, 0.05, 0.90]$\\ \hline
\label{tab:distributions}
\end{tabular} 
\end{minipage}
\end{minipage}
\caption{Search area and cell-wise environment distributions}
\label{fig:searchArea}
\end{figure}

Fig.~\ref{fig:searchArea} shows a search area that is partitioned into regions A1 through A5. For each region, the corresponding probability distribution $\Pi = [p_1, p_2, p_3 ]$ is given, where $p_j$ is the probability that the environment is $w_j$. For example, for the cells labeled A2, there is 0.15 probability that the environment is $w_1$, 0.2 probability that the environment is $w_2$, and 0.65 probability that the environment is $w_3$. The relative costs of over and underestimating the environmental conditions are $c_1 = 1$ and $c_2 = 3$ so that overestimation is penalized more than underestimation. The mission length is 60 for the search and search/environmental characterization vehicles and 35 for the characterization vehicle. We adopt a best-first branch-and-bound approach to compute the optimal paths~\cite{wah1985stochastic}.  

We consider two scenarios. In one scenario the search and the environmental characterization sensors operate on the same vehicle, and in the other scenario they operate on separate vehicles. When the sensors operate on separate vehicles, the objective of the search vehicle is to maximize anticipated estimation accuracy in~\eqref{eq:bestPath_searchonly}, and the objective of the characterization vehicle is to maximize the expected gain of characterization in~\eqref{eq:pathGain}. However, due to large computational requirements of computing~\eqref{eq:pathGain}, we instead approximate the solution of~\eqref{eq:pathGain} as described in Section~\ref{sec:approximation_cell}. When both sensors operate on the same vehicle, the objective of the vehicle is to maximize expected estimation accuracy in~\eqref{eq:expUtilPathCompound}. 

We define the \emph{error in search performance} after a mission as the difference between the actual estimation accuracy when the true environment is known and the anticipated estimation accuracy when the environment is uncertain. We use the error in search performance as a measure to evaluate the efficacy of the proposed approaches in each scenario, and show that the proposed approach yields smaller search performance error, which is predicted by our selection of cost function.  We also show that search performance (probability of correct estimate) increases modestly, although our approach does not directly seek to increase estimation accuracy.

\begin{figure*}[t!]
\centering
\begin{tabular}{l @{\hskip 0.5in} c @{\hskip 0.5in} r}
\subfloat[]{\includegraphics[width=0.26\textwidth,keepaspectratio,clip]{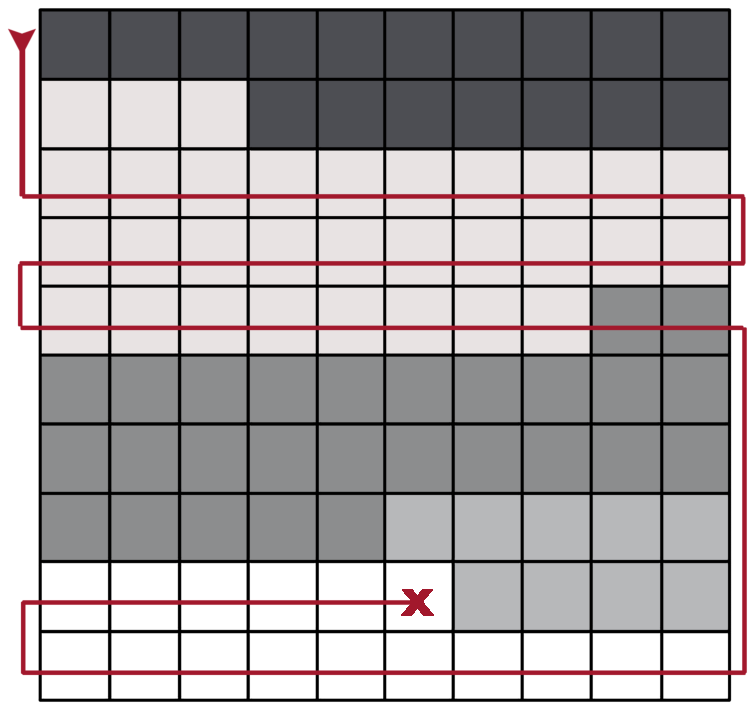}\label{fig:combinedM1}} &
\subfloat[]{\includegraphics[width=0.26\textwidth,keepaspectratio,clip]{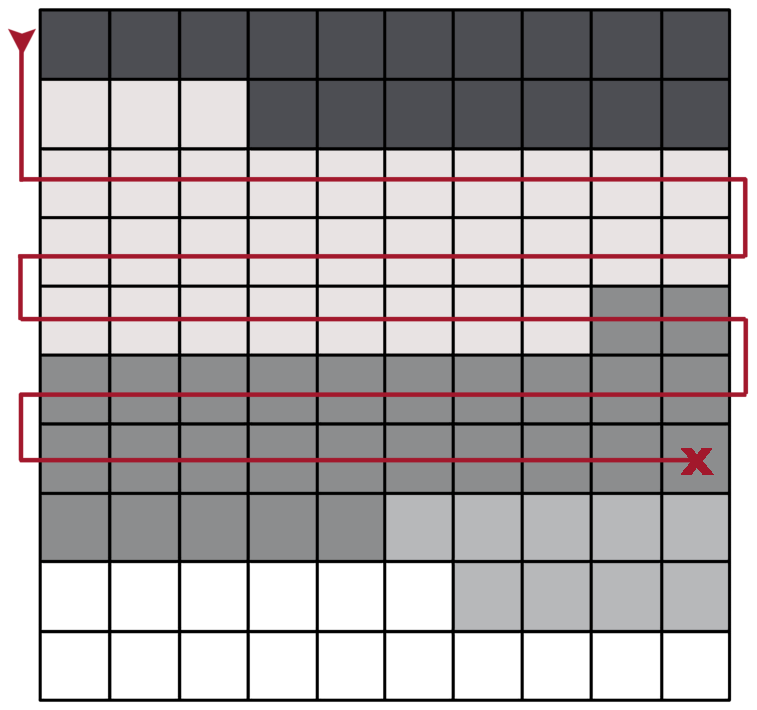}\label{fig:combinedM3}} &
\subfloat[]{\includegraphics[width=0.26\textwidth,keepaspectratio,clip]{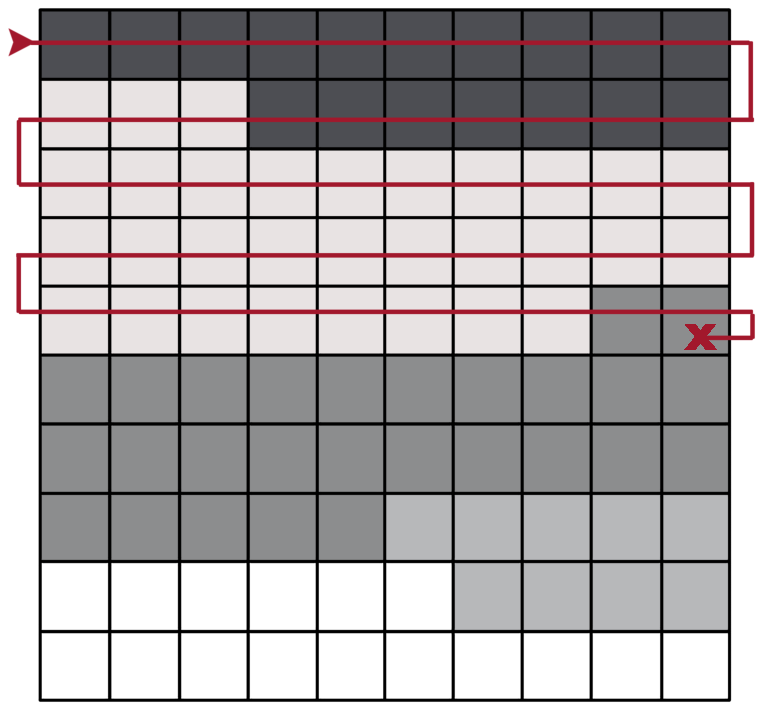}\label{fig:combined_mtl}} \\
\subfloat[]{\includegraphics[width=0.26\textwidth,keepaspectratio,clip]{path_search_proposed.eps}\label{fig:combinedM2}}& 
\subfloat[]{\includegraphics[width=0.26\textwidth,keepaspectratio,clip]{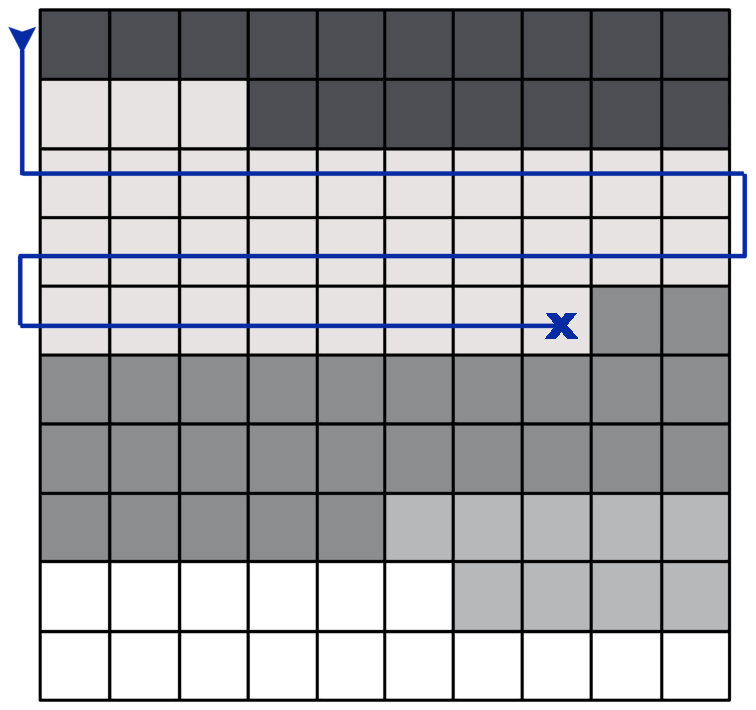}\label{fig:char_loss}}& 
\subfloat[]{\includegraphics[width=0.26\textwidth,keepaspectratio,clip]{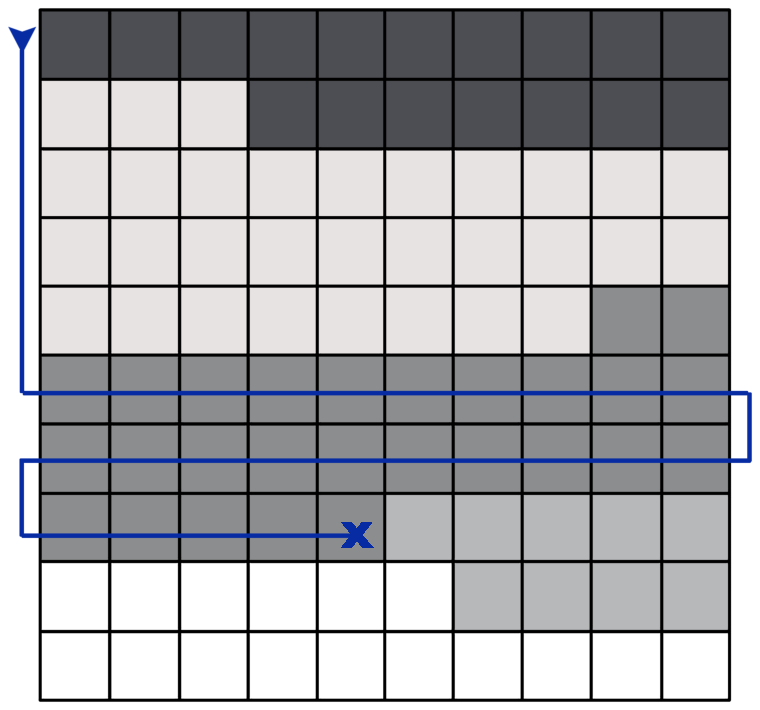}\label{fig:char_entropy}}
\end{tabular}
\caption{Optimal trajectories for search and characterization. Figures (a-c): trajectories for the case both sensors operate on the same vehicle when (a) proposed approach is employed (b) entropy change maximization method is employed, and (c) the mowing-the-lawn approach is employed. Figures (e-f): characterization vehicle's trajectories for the case the sensors operate on separate vehicles when the characterization locations are selected (e) by our proposed approach, (f) by entropy change maximization method. Figure (d) shows the search vehicle's trajectory when no environment information acquired.}
\label{fig:simulations}
\end{figure*}

When the sensors are on separate vehicles and characterization precedes search, we compare the proposed approximate approach in Section~\ref{sec:approximation_cell} with the entropy change maximization method described in Section~\ref{sec:entropy}. Fig.~\ref{fig:char_loss} shows the trajectory for the environmental characterization vehicle when using our proposed approximate approach in Section~\ref{sec:approximation_cell}, and Figure~\ref{fig:char_entropy} shows the trajectory when using the entropy change approach in Section~\ref{sec:entropy}.  Neither environmental characterization path visits A1 because the environments in those locations are completely known.  We note that the environmental characterization path in Figure~\ref{fig:char_loss} that was selected using our approach does not visit the most uncertain environments.   We find in practice that it tends to visit environments that are both uncertain \emph{and} likely to be where follow-on search missions will occur.

When both sensors operate on the same vehicle, we compare the proposed approach in~\eqref{eq:bestPath_searchonlyCompound} with entropy change maximization method and with a mowing-the-lawn approach. The latter arises often in subsea applications such as mine-hunting. We note that the entropy change maximization method described in Section~\ref{sec:entropy} accounts only for the entropy change of the environmental distributions. However, when both sensors are placed on the same vehicle, the vehicle acquires environmental measurement and search measurement simultaneously. Thus, we modify~\eqref{eq:characterizationpath_entropy} as

\begin{align}
\gamma^\star = \arg\max_{\gamma \in \Omega_\gamma} J_{\gamma}(X) + \beta J_{\gamma}(E) \nonumber
\end{align}

\noindent where $J(X)$ denotes the entropy change in $P(X)$, the number of objects, and $\beta$ is the relative weight of the entropy change in $P(E)$ compared to the entropy change in $P(X)$. Since the objective is to reduce the uncertainty in the number of objects, we choose $0 < \beta < 1$. Fig.~\ref{fig:combined_mtl} shows the mowing-the-lawn trajectory where the vehicle travels through the search area back and forth without planning the path until the mission length is met. Fig.~\ref{fig:combinedM1} shows the trajectory for the proposed approach and Fig.~\ref{fig:combinedM3} shows the trajectory for the entropy change maximization method with $\beta = 0.5$. We also compute the optimal search trajectory when there is no environmental characterization to show the value of acquiring environmental information. The corresponding trajectory for this case is shown in Fig.~\ref{fig:combinedM2}.


\begin{figure*}[t!]
\centering
\begin{tabular}{l r}
\subfloat[]{\includegraphics[width=0.5\textwidth,keepaspectratio,clip]{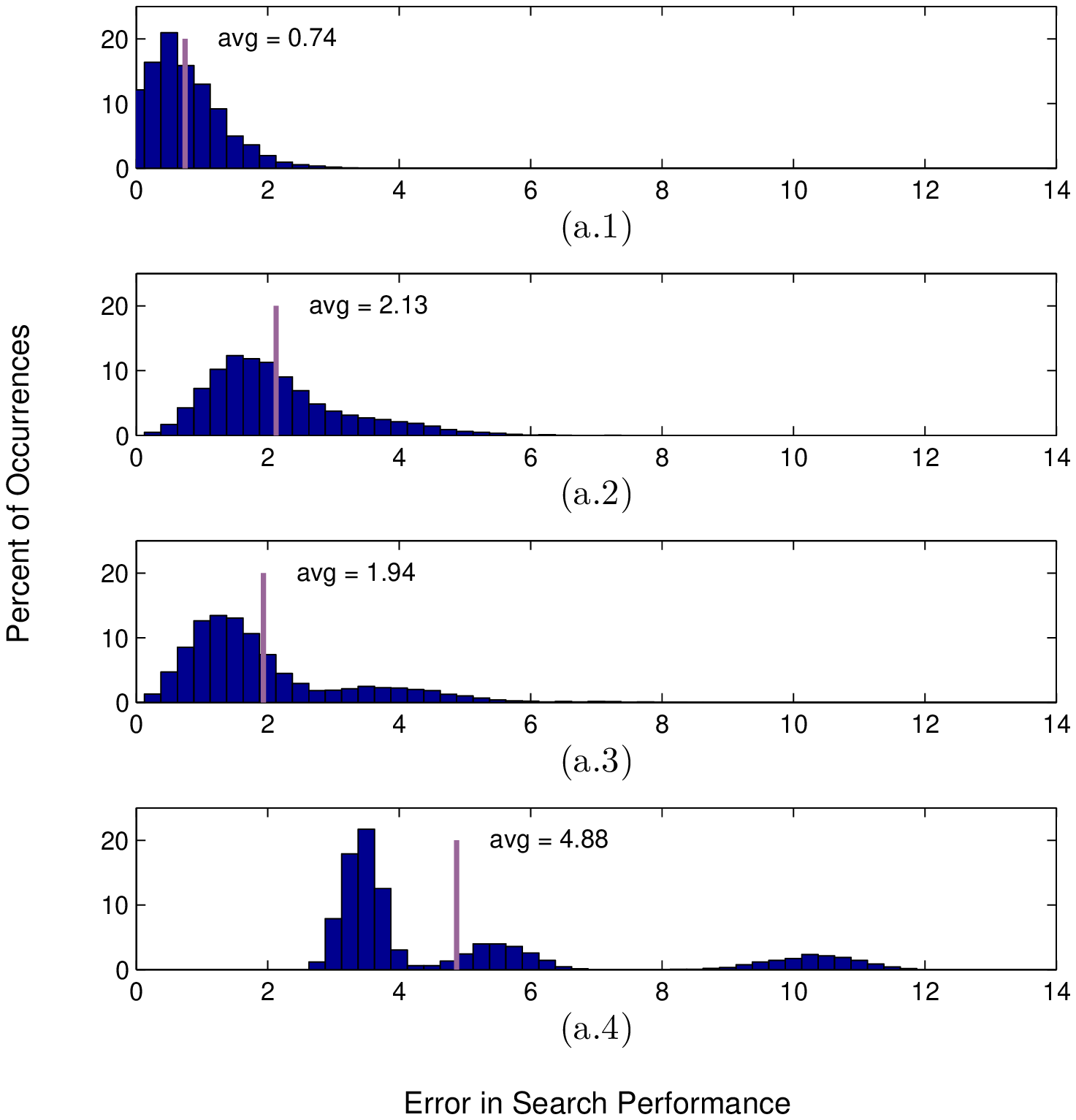}\label{fig:mc_combined_error}} &
\subfloat[]{\includegraphics[width=0.5\textwidth,keepaspectratio,clip]{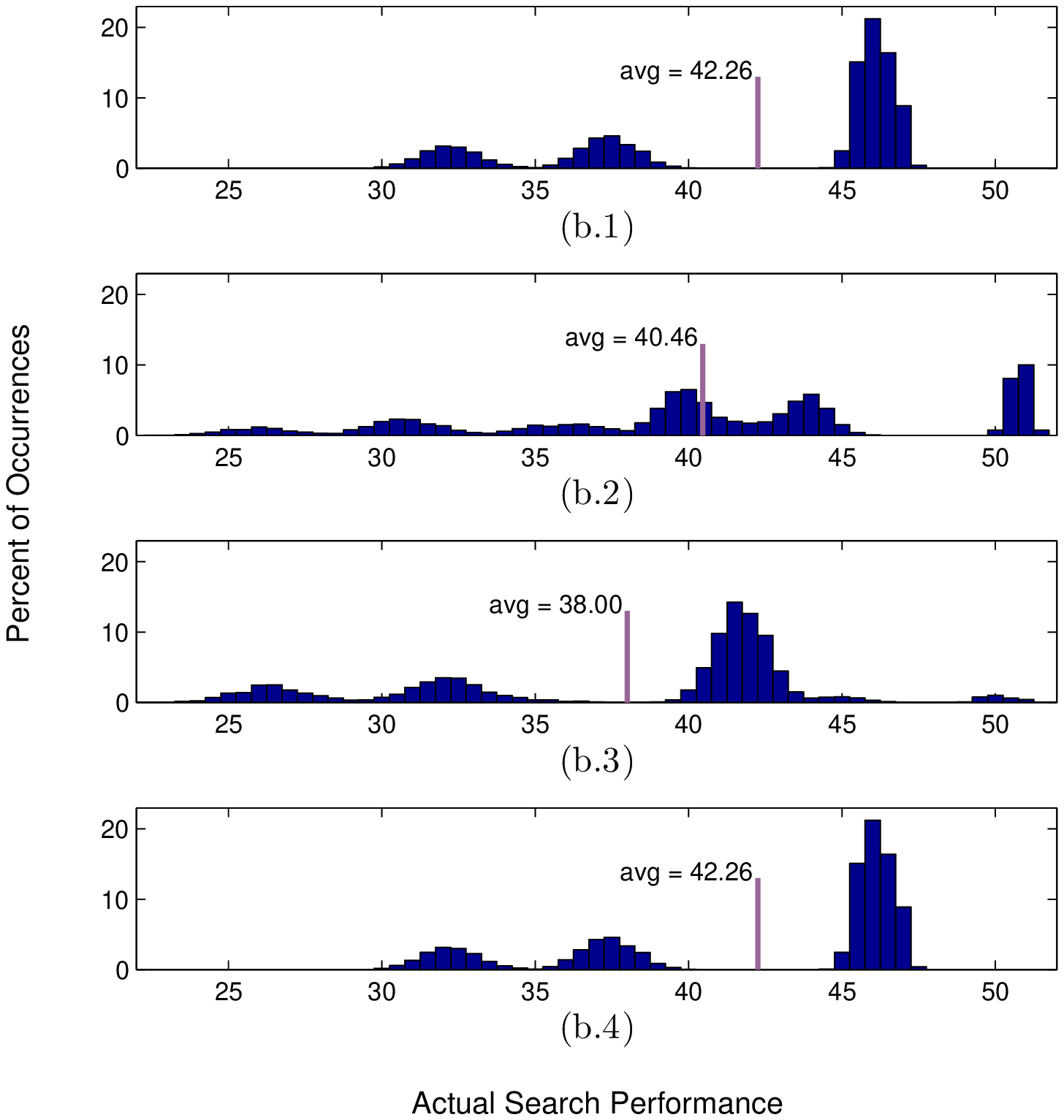}\label{fig:mc_combined_true}} 
\end{tabular}
\caption{Percent of occurrences for (a) error in search performance and (b) actual search performance when both sensors operate on \emph{the same vehicle}. From top to bottom, (a.1) and (b.1) correspond to the proposed approach, (a.2) and (b.2) correspond to the entropy change maximization method, (a.3) and (b.3) correspond to the mowing-the-lawn approach, and (a.4) and (b.4) correspond to the case where environment information is not available. Note that the horizontal axes is the negative log of the results. Smaller values for (a) imply less error in search performance and larger values for (b) imply better search performance.}
\label{fig:simulations_combined}
\end{figure*}

Search performance after a mission depends on the acquired observations during the mission. Thus, we conduct Monte Carlo simulations to assess the effects due to random nature of observations. For each cell in the search area, we randomly generate the true environment $e$ from the environmental distributions in Fig.~\ref{fig:searchArea} and the true number of objects $x$ from a uniform distribution. Assuming that a cell can be visited by a vehicle at most $k$ times, we randomly generate the set of search measurements $z$ and the set of environmental measurements $y$ from the sensor models $P(z \mid x, e)$ and $P(y \mid e)$ given true environment $e$ and true number of objects $x$. When a vehicle visits a location, it acquires randomly generated observation(s). For each test, we compute the anticipated search performance and the actual search performance. Note that the actual search performance can be computed since the true environment is assumed to be known. We then compute the error in search performance which is the difference between the anticipated search performance and the actual search performance. We show that the error in search performance is significantly reduced when our proposed approach is employed.

\subsection*{Both sensors operating simultaneously on a single vehicle}

Fig.~\ref{fig:simulations_combined} shows the results after 10000 iterations for the case both sensors operate on the same vehicle. Fig.~\ref{fig:mc_combined_error} on the left is the percent of occurrences of the error in search performance, and Fig.~\ref{fig:mc_combined_true} on the right is the percent of occurrences of the actual search performance. For convenience, we compute the actual search performance after traversing an optimal path $\gamma$ and acquiring the search measurements $z_{\gamma}$ as

\begin{flalign*}
- \Big ( \log & \big ( \prod_{i \in \mathcal{G} } \max_{x_i} P( x_i ) \big ) \\ 
& - \log \big (  \prod_{i \in \gamma} \max_{x_i} P( x_i \mid z_i, e_i ) \times \prod_{i \in \mathcal{G} \setminus \gamma} \max_{x_i} P( x_i ) \big )  \Big )
\end{flalign*}

\noindent where $e_i$ is the actual environment in cell $i$. That is, the actual search performance is the difference between the prior certainty in the number of objects before acquiring any measurement and the posterior certainty in the number of objects after acquiring the search measurements along the path. Loosely speaking, the actual search performance plotted in Fig.~\ref{fig:mc_combined_true} represents the amount of information we acquire on the number of objects after traversing the corresponding optimal search path. Thus, smaller values for Fig.~\ref{fig:mc_combined_error} imply less error in search performance and larger values for Fig.~\ref{fig:mc_combined_true} imply better search performance. The displayed results are the negative log of the computed search performance. The subplots from top to bottom are the results when 1) our proposed approach is employed, 2) the entropy change maximization method is employed, 3) the mowing-the-lawn approach is employed, and 4) environmental information is not available so that the vehicle acquires only the search measurements. The average value of results for each test is also shown in the plots. The simulations show that

\begin{itemize}
\item The proposed approach yields smaller error in search performance compared to the entropy change maximization and mowing-the-lawn. With respect to the case where there is no environment information (in Fig.~\ref{fig:mc_combined_error}.4), our proposed approach achieves 85\% error reduction while entropy change maximization achieves 56\% and mowing-the-lawn achieves 60\%, on average. In addition, the actual search performance when using our approach is no worse than the actual search performance when using the other methods. 

\item Fig.~\ref{fig:mc_combined_error}.1 shows that in many of the iterations, the error in search performance is very close to zero. This implies that, in these trials, we correctly estimate the environmental conditions in each visited cell. We note that this is also due to the sensor model we choose in~\eqref{eq:sensormodel_environment} for environment characterization.        

\item The average error for the mowing-the-lawn approach is smaller than the average error for entropy change maximization method. This is because mowing-the-lawn approach visits A1 that has no uncertainty in the environment while the entropy change maximization method visits A3 where the environmental uncertainty is greatest. However, as the environment in A1 is the least informative, the average actual search performance for mowing-the-lawn approach is the worst among all methods.   

\item Note that Fig.~\ref{fig:mc_combined_true}.1 and Fig.~\ref{fig:mc_combined_true}.4 are identical. This is because the search locations selected by the proposed approach (in Fig.~\ref{fig:combinedM1}) are the same locations selected when environment information is not available (in Fig.~\ref{fig:combinedM2}) for given search area characteristics. However, the anticipated search performances for these two cases are different. Indeed, comparing Fig.~\ref{fig:mc_combined_error}.1 with Fig.~\ref{fig:mc_combined_error}.4 shows that the anticipated search performance when environment information is available is significantly more accurate than the anticipated search performance when there is no environment information. Hence, a benefit of characterizing the environment is to better anticipate the true search performance.     

\end{itemize}

\subsection*{Each sensor on separate vehicles}

\begin{figure*}[t!]
\centering
\begin{tabular}{l r}
\subfloat[]{\includegraphics[width=0.5\textwidth,keepaspectratio,clip]{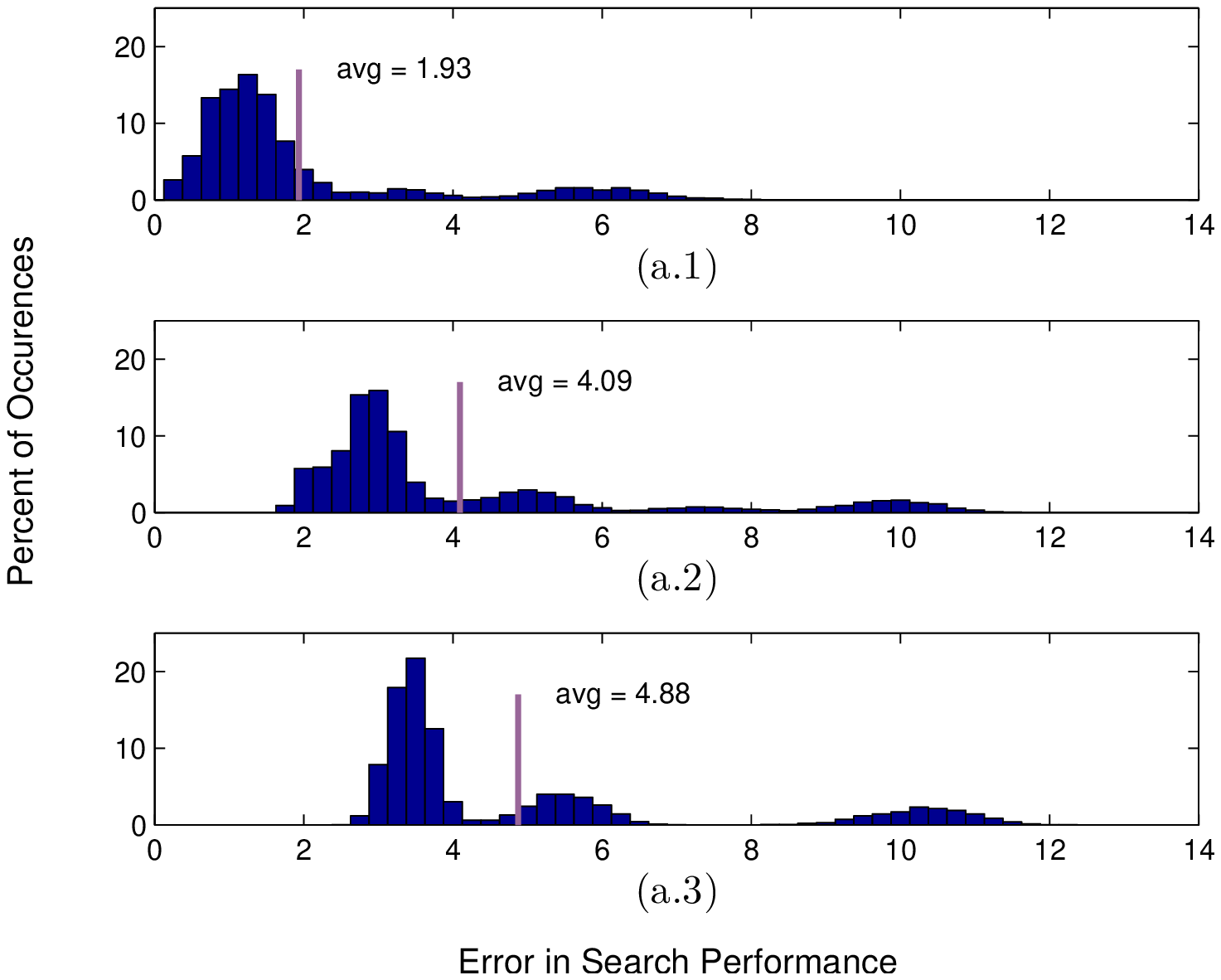}\label{fig:mc_characterization_error}} &
\subfloat[]{\includegraphics[width=0.5\textwidth,keepaspectratio,clip]{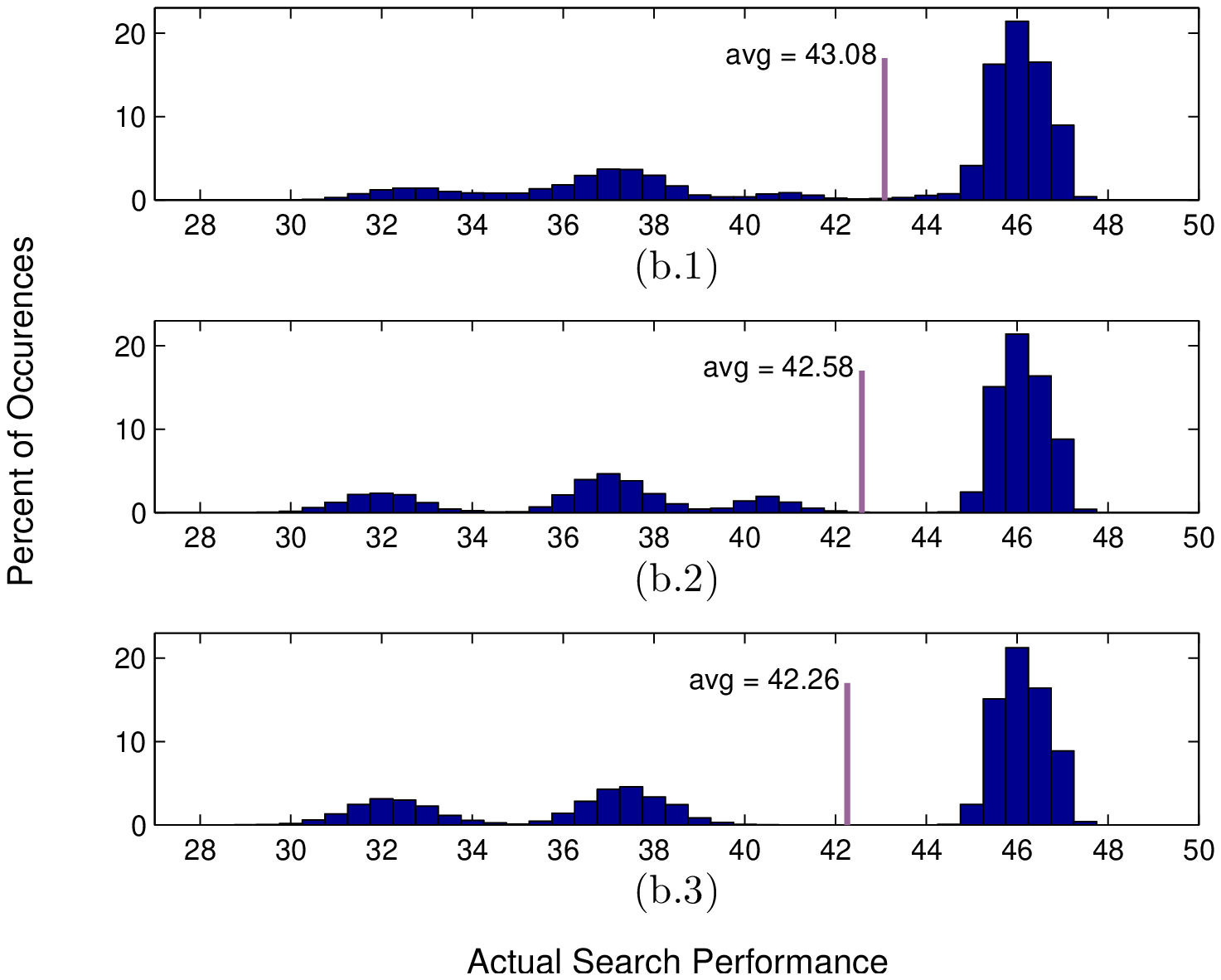}\label{fig:mc_characterization_true}}
\end{tabular}
\caption{Percent of occurrences for (a) error in search performance and (b) actual search performance when search and characterization are performed on \emph{separate vehicles}. From top to bottom, (a.1) and (b.1) correspond to the proposed approach, (a.2) and (b.2) correspond to the entropy change maximization method, and (a.3) and (b.3) correspond to the case where environment information is not available. Note that the horizontal axes is the negative log of the results. Smaller values for (a) imply less error in search performance and larger values for (b) imply better search performance.}
\label{fig:simulations_characterization}
\end{figure*}

The results when search and environmental characterization tasks are performed on separate vehicles are shown in Fig.~\ref{fig:simulations_characterization}. Again, the left plot is the percent of occurrences of the error in search performance, and the right plot is the percent of occurrences of actual search performance. The subplots from top to bottom are the results when 1) the locations that yield the greatest reduction of uncertainty in search performance are characterized, 2) the locations that maximize the entropy change are characterized, and 3) there is no environmental characterization and the search vehicle plans its path by using the prior environmental distributions. We note that Fig.~\ref{fig:mc_characterization_error}.3 and Fig.~\ref{fig:mc_characterization_true}.3 are the same plots given in Fig.~\ref{fig:mc_combined_error}.4 and Fig.~\ref{fig:mc_combined_true}.4, and we show them here for convenience of comparison. It is seen that

\begin{itemize}
\item The average error is significantly smaller when environmental characterization is performed at the locations selected by our proposed approach. With respect to the case where there is no environment information (in Fig.~\ref{fig:mc_characterization_error}.3), our proposed approach achieves 60\% error reduction while entropy change maximization achieves only 16\%, on average. Note that the distribution in Fig.~\ref{fig:mc_characterization_error}.2 is very similar to the distribution in Fig.~\ref{fig:mc_characterization_error}.3. This is because entropy change maximization fails to improve the performance of a follow-on search mission since it leads the vehicle to explore the parts of the search area that are less likely to be searched. 

\item The average error when the sensors are on different vehicles is higher than when both sensors operate on the same vehicle since the search vehicle may search the locations that are not characterized. On the other hand, this results in average actual search performance to be better since the search vehicle can skip the locations that are characterized and found to be uninteresting for search. 
      
\end{itemize}       

The results of Monte Carlo simulations show that our proposed approaches to select the characterization locations outperform the other strategies that frequently exist in the literature.    

\section{Conclusions}

In this paper, we address the case where environmental information can be acquired to improve the performance of a search mission. We consider different scenarios where the search sensor and the environmental characterization sensor can be placed on the same vehicle or on separate vehicles. For each scenario, we derive a decision-theoretic cost function to compute the locations where environmental information should be acquired. We show that when the search sensor and the environmental characterization sensor are placed on separate vehicles, environmental information should be acquired at the locations where the greatest reduction of the uncertainty in anticipated estimation accuracy will occur. For the case where the search sensor and the environmental characterization sensor are placed on the same vehicle, we show that the expected estimation accuracy should be maximized. The results of  the numerical illustrations show that for each scenario, our proposed approaches yield smaller error in search performance.      

\section*{Acknowledgements}
The authors gratefully acknowledge the support of the Office of Naval Research via grants N00014-12-1-0055 and N00014-16-1-2092. The assistance provided by Dr. Hongxiao Zhu (Department of Statistics, Virginia Tech) is greatly appreciated.

\bibliographystyle{IEEEtran}

\bibliography{harun.search}

\appendix

\subsection{Table showing the list of variables}
\label{sec:appendix_variables}

\begin{table}[h!]
\centering
\caption{A list of variables}
\begin{tabular}{l l}  
\toprule
{\bf Variable}  &  {\bf Description} \\
\midrule
$X$ 			& \makecell{random variable denoting the number of targets in a cell}\\
$E$				& \makecell{random variable denoting environmental conditions in a cell} \\
$Z$				& \makecell{the set of search measurements and also \\ random variable denoting a search measurement} \\
$Y$				& \makecell{the set of environment measurements and also \\ random variable denoting an environment measurement} \\
$x_i$			& \makecell{true number of objects in cell $i$} \\
$e_i$			& \makecell{true environment in cell $i$} \\
$z_i$  			& \makecell{search measurement(s) acquired from cell $i$} \\
$y_i$			& \makecell{environment measurement(s) acquired from cell $i$} \\
$\delta_X(\cdot)$	& \makecell{an estimate of the number of objects} \\
$\delta_E(\cdot)$	& \makecell{an estimate of the environment} \\
$\mathcal{G}$  	& search grid    \\
$n_r, n_c$      & \makecell{number of rows and columns, respectively} \\
$w_j$			& \makecell{environment type $j$} \\
$\gamma$			& \makecell{a candidate path for the search vehicle} \\
$\eta$				& \makecell{a candidate path for the characterization vehicle} \\
$\Omega$	& \makecell{the set of candidate paths} \\
$\Cgamma$	& \makecell{budget constraint on search mission} \\
$\Ceta$	& \makecell{budget constraint on environment characterization mission} \\
$\mathcal{C}(\cdot)$	& \makecell{cost of traversing a path} \\
\bottomrule
\end{tabular}
\label{tab:variables}
\end{table}

\subsection{Proof of Theorem~\ref{theorem:bound_on_gain}}
\label{sec:appendix_proof}

First, observe that when $c_1, c_2 \leq 1$ in~\eqref{eq:lossFunction} 

\begin{eqnarray*}
\mathcal{R}(y) & = & \E{ L_V\loss{ e, \delta_E^\star } } - \E{ L_V\loss{ e, \delta_E^\star(y) } \mid y \in Y } \\
& \leq & \E{ L_V\loss{ e, \delta_E^\star } } \\
& \leq & \max_{w_i, w_j} \big ( \V{w_i} - \V{w_j} \big ) \\
& \leq & 1
\end{eqnarray*} 

\noindent for all $y \in Y$.

The difference between $\Es{G_\eta}$ and $\Es{\hat{G}_\eta}$ is

\begin{align*}
& \big | \Es{G_\eta} - \Es{\hat{G}_\eta} \big | \\
\begin{split}
&  = \Big | \sum\limits_{q_i \in \eta } \sum\limits_{y_{q_i} } \Big ( \P{ Y_{q_i} = y_{q_i} } \mathcal{R}(y_{q_i}) \Big ) \times \bar{\mathbf{P}}_{\eta \setminus q_i} \\
& \qquad \qquad -  \sum\limits_{q_i \in \eta } \sum\limits_{y_{q_i} } \Big ( \P{ Y_{q_i} = y_{q_i} } \mathcal{R}(y_{q_i}) \Big ) \times \hat{\mathbf{P}}_{\eta \setminus q_i} \Big |
\end{split} \\
& = \Big | \sum\limits_{q_i \in \eta } \sum\limits_{y_{q_i} } \Big ( \P{ Y_{q_i} = y_{q_i} } \mathcal{R}(y_{q_i}) \Big ) \times ( \bar{\mathbf{P}}_{\eta \setminus q_i} - \hat{\mathbf{P}}_{\eta \setminus q_i} ) \Big | \\
& \leq  \sum\limits_{q_i \in \eta } \sum\limits_{y_{q_i} } \Big ( \P{ Y_{q_i} = y_{q_i} } \mathcal{R}(y_{q_i}) \Big ) \times \big | \bar{\mathbf{P}}_{\eta \setminus q_i} - \hat{\mathbf{P}}_{\eta \setminus q_i} \big | \\
& \leq  \sum\limits_{q_i \in \eta } \sum\limits_{y_{q_i} } \P{ Y_{q_i} = y_{q_i} }  \times \big | \bar{\mathbf{P}}_{\eta \setminus q_i} - \hat{\mathbf{P}}_{\eta \setminus q_i} \big | \\
& =  \sum\limits_{q_i \in \eta }  \big | \bar{\mathbf{P}}_{\eta \setminus q_i} - \hat{\mathbf{P}}_{\eta \setminus q_i} \big | \\
& =  |\eta|  \big | \bar{\mathbf{P}}_{\eta \setminus q_i} - \hat{\mathbf{P}}_{\eta \setminus q_i} \big | \\
\end{align*} 

Due to the bound on the difference between $\bar{\mathbf{P}}_{\eta \setminus q_i}$ and $\hat{\mathbf{P}}_{\eta \setminus q_i}$  in~\eqref{eq:Pbar}, it follows that

\begin{align*}
\P{ \big | \Es{G_\eta} - \Es{\hat{G}_\eta} & \big | < \epsilon \bar{\mathbf{N}} |\eta| } \\
& = \P{ |\eta| | \bar{\mathbf{P}}_{\eta \setminus q_i} - \hat{\mathbf{P}}_{\eta \setminus q_i} | < \epsilon \bar{\mathbf{N}} |\eta| } \\
& = \P{ | \bar{\mathbf{P}}_{\eta \setminus q_i} - \hat{\mathbf{P}}_{\eta \setminus q_i} | < \epsilon \bar{\mathbf{N}} } \\ 
& \geq 1 - 2 \exp ^{-2\epsilon^2 \bar{\mathbf{N}} }
\end{align*}

\subsection{Proof of Corollary~\ref{theorem:bound_on_optimalgain}}
\label{sec:appendix_corollary}

By Theorem~\ref{theorem:bound_on_gain}, we obtain the bounds for $\eta^\star$ in~\eqref{eq:bestPath_searchonlyAfterY} and for $\hat{\eta}$ in~\eqref{eq:approximate_optimalpath}

\begin{eqnarray*}
\P{ \big | \Es{G_{\eta^\star}} - \Es{\hat{G}_{\eta^\star}} \big | < \epsilon \bar{\mathbf{N}} |\eta^\star| }  & \geq & 1 - 2 \exp ^{-2\epsilon^2 \bar{\mathbf{N}} } \\
\P{ \big | \Es{G_{\hat{\eta}}} - \Es{\hat{G}_{\hat{\eta}}} \big | < \epsilon \bar{\mathbf{N}} |\hat{\eta}| }  & \geq & 1 - 2 \exp ^{-2\epsilon^2 \bar{\mathbf{N}} }
\end{eqnarray*}

Note that since $\hat{\eta}$ maximizes~\eqref{eq:approximate_optimalpath}, $\Es{\hat{G}_{\hat{\eta}}} \geq \Es{\hat{G}_{\eta^\star}}$. Hence, 

\begin{align*}
\Es{G_{\eta^\star}} - \Es{& G_{\hat{\eta}}} =  \Es{G_{\eta^\star}} - \Es{\hat{G}_{\eta^\star}} + \Es{\hat{G}_{\eta^\star}} - \Es{G_{\hat{\eta}}} \\
& \leq  \big ( \Es{G_{\eta^\star}} - \Es{\hat{G}_{\eta^\star}} \big ) + \big ( \Es{\hat{G}_{\hat{\eta}}} - \Es{G_{\hat{\eta}}} \big ) \\
& \leq  \big | \Es{G_{\eta^\star}} - \Es{\hat{G}_{\eta^\star}} \big | + \big | \Es{\hat{G}_{\hat{\eta}}} - \Es{G_{\hat{\eta}}} \big |
\end{align*}

Then, assuming $|\eta^\star| = |\hat{\eta}| = |\eta|$, error in approximation of the optimal characterization gain is

\begin{align*}
& \P{ \Es{G_{\eta^\star}} - \Es{G_{\hat{\eta}}}  < 2 \epsilon \bar{\mathbf{N}} |\eta| } \\
& \geq \P{ \big | \Es{G_{\eta^\star}} - \Es{\hat{G}_{\eta^\star}} \big | + \big | \Es{\hat{G}_{\hat{\eta}}} - \Es{G_{\hat{\eta}}} \big |  < 2 \epsilon \bar{\mathbf{N}} |\eta| } \\
& \geq 1 - 2 \exp ^{-2\epsilon^2 \bar{\mathbf{N}} }
\end{align*}

\end{document}